\PassOptionsToPackage{unicode}{hyperref}
\PassOptionsToPackage{hyphens}{url}
\PassOptionsToPackage{dvipsnames,svgnames,x11names}{xcolor}
\documentclass[
]{article}

\usepackage{amsmath,amssymb}
\usepackage{iftex}
\ifPDFTeX
  \usepackage[T1]{fontenc}
  \usepackage[utf8]{inputenc}
  \usepackage{textcomp} 
\else 
  \usepackage{unicode-math}
  \defaultfontfeatures{Scale=MatchLowercase}
  \defaultfontfeatures[\rmfamily]{Ligatures=TeX,Scale=1}
\fi
\usepackage{lmodern}
\ifPDFTeX\else  
\fi
\IfFileExists{upquote.sty}{\usepackage{upquote}}{}
\IfFileExists{microtype.sty}{
  \usepackage[]{microtype}
  \UseMicrotypeSet[protrusion]{basicmath} 
}{}
\makeatletter
\@ifundefined{KOMAClassName}{
  \IfFileExists{parskip.sty}{%
    \usepackage{parskip}
  }{
    \setlength{\parindent}{0pt}
    \setlength{\parskip}{6pt plus 2pt minus 1pt}}
}{
  \KOMAoptions{parskip=half}}
\makeatother
\usepackage{xcolor}
\ifx\paragraph\undefined\else
  \let\oldparagraph\paragraph
  \renewcommand{\paragraph}[1]{\oldparagraph{#1}\mbox{}}
\fi
\ifx\subparagraph\undefined\else
  \let\oldsubparagraph\subparagraph
  \renewcommand{\subparagraph}[1]{\oldsubparagraph{#1}\mbox{}}
\fi

\usepackage{longtable,booktabs,array}
\usepackage{calc} 
\usepackage{etoolbox}
\makeatletter
\patchcmd\longtable{\par}{\if@noskipsec\mbox{}\fi\par}{}{}
\makeatother
\IfFileExists{footnotehyper.sty}{\usepackage{footnotehyper}}{\usepackage{footnote}}
\makesavenoteenv{longtable}
\usepackage{graphicx}
\makeatletter
\def\maxwidth{\ifdim\Gin@nat@width>\linewidth\linewidth\else\Gin@nat@width\fi}
\def\maxheight{\ifdim\Gin@nat@height>\textheight\textheight\else\Gin@nat@height\fi}
\makeatother
\setkeys{Gin}{width=\maxwidth,height=\maxheight,keepaspectratio}
\makeatletter
\def\fps@figure{htbp}
\makeatother
\newlength{\cslhangindent}
\newlength{\csllabelwidth}
\newlength{\cslentryspacingunit} 
\newenvironment{CSLReferences}[2] 
 {
  \setlength{\parindent}{0pt}
  \ifodd #1
  \let\oldpar\par
  \def\par{\hangindent=\cslhangindent\oldpar}
  \fi
  \setlength{\parskip}{#2\cslentryspacingunit}
 }%
 {}
\usepackage{calc}

\usepackage{arxiv}
\usepackage{orcidlink}
\usepackage{amsmath}
\usepackage[T1]{fontenc}
\makeatletter
\@ifpackageloaded{caption}{}{\usepackage{caption}}
\AtBeginDocument{%
\ifdefined\contentsname
  \renewcommand*\contentsname{Table of contents}
\else
  \newcommand\contentsname{Table of contents}
\fi
\ifdefined\listfigurename
  \renewcommand*\listfigurename{List of Figures}
\else
  \newcommand\listfigurename{List of Figures}
\fi
\ifdefined\listtablename
  \renewcommand*\listtablename{List of Tables}
\else
  \newcommand\listtablename{List of Tables}
\fi
\ifdefined\figurename
  \renewcommand*\figurename{Figure}
\else
  \newcommand\figurename{Figure}
\fi
\ifdefined\tablename
  \renewcommand*\tablename{Table}
\else
  \newcommand\tablename{Table}
\fi
}
\@ifpackageloaded{float}{}{\usepackage{float}}
\floatstyle{ruled}
\@ifundefined{c@chapter}{\newfloat{codelisting}{h}{lop}}{\newfloat{codelisting}{h}{lop}[chapter]}
\floatname{codelisting}{Listing}

\makeatother
\makeatletter
\@ifpackageloaded{caption}{}{\usepackage{caption}}
\@ifpackageloaded{subcaption}{}{\usepackage{subcaption}}
\makeatother
\makeatletter
\@ifpackageloaded{tcolorbox}{}{\usepackage[skins,breakable]{tcolorbox}}
\makeatother
\makeatletter
\@ifundefined{shadecolor}{\definecolor{shadecolor}{rgb}{.97, .97, .97}}
\makeatother
\ifLuaTeX
  \usepackage{selnolig}  
\fi
\IfFileExists{bookmark.sty}{\usepackage{bookmark}}{\usepackage{hyperref}}
\IfFileExists{xurl.sty}{\usepackage{xurl}}{} 
\hypersetup{
  pdftitle={Convolutional Neural Networks for Neuroimaging in Parkinson's Disease: Is Preprocessing Needed?},
  pdfauthor={Juan M. Górriz; Javier Ramírez; Andrés Ortiz},
  colorlinks=true,
  linkcolor={blue},
  filecolor={Maroon},
  citecolor={Blue},
  urlcolor={Blue},
  pdfcreator={LaTeX via pandoc}}

\renewcommand{\today}{12/09/2018}
\newcommand{\runninghead}{A Preprint }
\renewcommand{\runninghead}{Electronic version of an article published
as INT J NEURAL SYST 28 (10), 2018, 1850035
{[}10.1142/S0129065718500351{]} \copyright World Scientific Publishing
Company }
\title{Convolutional Neural Networks for Neuroimaging in Parkinson's
Disease: Is Preprocessing Needed?}
\def\asep{\\\\\\ } 
\author{\textbf{Francisco J.
Martinez-Murcia}~\orcidlink{0000-0001-8146-7056}\\\\Dpt. of Signal
Theory, Networking and Communications. University of Granada. Granada,
Spain\\\\\href{mailto:fjesusmartinez@ugr.es}{fjesusmartinez@ugr.es}\asep\textbf{Juan
M. Górriz}\\\\Dpt. of Signal Theory, Networking and Communications.
University of Granada. Granada, Spain\\\\\asep\textbf{Javier
Ramírez}\\\\Dpt. of Signal Theory, Networking and Communications.
University of Granada. Granada, Spain\\\\\asep\textbf{Andrés
Ortiz}\\\\Department of Communications Engineering. University of
Malaga. Malaga, Spain\\\\}
\date{12/09/2018}
\begin{document}
\maketitle
\begin{abstract}
Spatial and intensity normalization are nowadays a prerequisite for
neuroimaging analysis. Influenced by voxel-wise and other univariate
comparisons, where these corrections are key, they are commonly applied
to any type of analysis and imaging modalities. Nuclear imaging
modalities such as PET-FDG or FP-CIT SPECT, a common modality used in
Parkinson's Disease diagnosis, are especially dependent on intensity
normalization. However, these steps are computationally expensive and
furthermore, they may introduce deformations in the images, altering the
information contained in them. Convolutional Neural Networks (CNNs), for
their part, introduce position invariance to pattern recognition, and
have been proven to classify objects regardless of their orientation,
size, angle, etc. Therefore, a question arises: how well can CNNs
account for spatial and intensity differences when analysing nuclear
brain imaging? Are spatial and intensity normalization still needed? To
answer this question, we have trained four different CNN models based on
well-established architectures, using or not different spatial and
intensity normalization preprocessing. The results show that a
sufficiently complex model such as our three-dimensional version of the
ALEXNET can effectively account for spatial differences, achieving a
diagnosis accuracy of 94.1\% with an area under the ROC curve of 0.984.
The visualization of the differences via saliency maps shows that these
models are correctly finding patterns that match those found in the
literature, without the need of applying any complex spatial
normalization procedure. However, the intensity normalization -and its
type- is revealed as very influential in the results and accuracy of the
trained model, and therefore must be well accounted.
\end{abstract}
\ifdefined\Shaded\renewenvironment{Shaded}{\begin{tcolorbox}[frame hidden, interior hidden, enhanced, borderline west={3pt}{0pt}{shadecolor}, sharp corners, breakable, boxrule=0pt]}{\end{tcolorbox}}\fi

\hypertarget{introduction}{%
\section{Introduction}\label{introduction}}

Deep learning (LeCun, Bengio, and Hinton 2015) is currently a major
trend in every research field. Its flexibility, adaptability and its
unprecedented accuracy in image classification has motivated its
adoption in many disciplines ranging from speech (Kim, Kim, and Seo
2004; Hinton et al. 2012) or pattern recognition (Garrido et al. 2016)
to drug discovery (Gawehn, Hiss, and Schneider 2016) and genomics
(Alipanahi et al. 2015). It is also gradually opening up in medical
image analysis (Liao et al. 2013; Olson and Perry 2013; Xu et al. 2014;
Greenspan, Ginneken, and Summers 2016; Ortiz et al. 2016; Martín-López
et al. 2017), and particularly in neuroimaging (Martín-López et al.
2017; Ortiz et al. 2016; Olson and Perry 2013), where large amounts of
data are analysed through well-established processes.

Large-scale analysis of neuroimaging such as the Statistical Parametric
Mapping (SPM) (Friston et al. 2007) or many other multivariate learning
techniques (I. A. Illán et al., n.d.; Fermín Segovia et al. 2012; Saxena
et al. 1998; Gorriz et al. 2017) require a multi-step preprocessing of
brain images. These steps include: motion correction, realignment,
linear or non-linear registration to a common space (frequently the MNI
space), segmentation, smoothing, intensity normalization, etc (F. J.
Martinez-Murcia, Górriz, and Ramírez 2016). Of these, non-linear
registration to a standard template -also known as spatial
normalization- is present in the vast majority of imaging tools, under
the assumption that the same anatomical positions must lay in the same
spatial coordinate in order to perform the analysis. This is true for
Computer Aided Diagnosis (CAD) systems using more traditional machine
learning algorithms such as Principal Component Analysis (PCA) (López et
al. 2011), or the Support Vector Machine (SVM) (I. A. Illán et al.,
n.d.; Fermín Segovia et al. 2012; Saxena et al. 1998; Gorriz et al.
2017; F. Martínez-Murcia et al. 2014) and Random Forests (Ramírez et al.
2010) classifiers. However, non-linear spatial normalization introduces
local deformations in the shape and size of different regions that may
lead to artifacts introduced by the anatomic standardization process
(Ishii et al. 2001; Reig et al. 2007; Martino et al. 2013).

Two major categories of neuroimaging can be defined: structural and
functional, depending on the information contained within. In the case
of functional imaging, intensity normalization is key. It helps to
perform a straightforward comparison between intensity level in each
image, despite other individual factors such as drug uptake, acquisition
time or scanner sensitivity (D. Salas-Gonzalez et al. 2012; I. A. Illán
et al., n.d.; Saxena et al. 1998; F. Martínez-Murcia et al. 2014). It
can be compared to feature standardization, a fundamental preprocessing
in machine learning, especially with learners such as Support Vector
Machines or Artificial Neural Networks (ANNs). However, while feature
standardization frequently converts each feature in all subjects to be
distributed with zero mean and unit variance, intensity normalization
changes the intensity levels of each subject independently, according to
different parameters such as global or maximum intensity.

Spatial and intensity normalization have been also applied as a part of
the preprocessing pipeline in applications of deep learning to medical
imaging (Yuvaraj et al. 2016; Francisco Jesús Martinez-Murcia et al.
2017; Hirschauer, Adeli, and Buford 2015). However, deep learning, and
in particular Convolutional Neural Networks (CNN) have shown great
ability in classifying objects into images regardless of their
orientation, size, angle, etc (Morabito et al. 2017; Koziarski and
Cyganek 2017; Ortega-Zamorano et al. 2017; Lin, Nie, and Ma 2017; Zhang
et al. 2017; Cha, Choi, and Büyüköztürk 2017; Rafiei and Adeli 2015,
2017, 2018; Rafiei et al. 2017; Acharya et al. 2017). Then, a question
arises: is spatial and/or intensity normalization needed in neuroimaging
applications of deep learning?

In this work we explore the effect of spatial and intensity
normalization of neuroimaging in the classification of Parkinson's
Disease (PD) patients from the well-established Parkinson's Progression
Markers Initiative (PPMI), using FP-CIT SPECT images. In
Section~\ref{sec-methods} we present the methodology followed, including
a description of the dataset, the spatial and intensity normalization
procedures evaluated, a description of our CNN, the data augmentation
procedures used and an introduction to the evaluation metrics. In
Section~\ref{sec-results}, the results of this evaluation are presented,
and further discussed at Section~\ref{sec-discussion}. Finally, in
Section~\ref{sec-conclussion}, some conclusions are drawn.

\hypertarget{sec-methods}{%
\section{Methodology}\label{sec-methods}}

\hypertarget{dataset}{%
\subsection{Dataset}\label{dataset}}

Data used in the preparation of this article were obtained from the
Parkinson's Progression Markers Initiative (PPMI) database
(\url{www.ppmi-info.org/data}). For up-to-date information on the study,
visit \url{www.ppmi-info.org}. The images in this database were imaged 4
+ 0.5 hours after the injection of between 111 and 185 MBq of DaTSCAN.
Raw projection data are acquired into a \(128 \times 128\) matrix
stepping each 3 degrees for a total of 120 projection into two 20\%
symmetric photopeak windows centered on 159 KeV and 122 KeV with a total
scan duration of approximately 30 - 45 minutes (Initiative 2010).

A total of \(N=642\) DaTSCAN images from this database were used in the
preparation of the article. Specifically, the baseline acquisition from
\(448\) subjects suffering from PD and \(194\) normal controls was used.
The images, roughly realigned, were then linearly resampled down to a
final size of (57,69,57), the input size of the network. Unless stated,
these are considered the `original', non-normalized images. For more
details on the demographics of this dataset, please check
Table~\ref{tbl-demographics}.

\hypertarget{tbl-demographics}{}
\begin{longtable}[]{@{}
  >{\raggedright\arraybackslash}p{(\columnwidth - 6\tabcolsep) * \real{0.1389}}
  >{\raggedright\arraybackslash}p{(\columnwidth - 6\tabcolsep) * \real{0.0833}}
  >{\raggedleft\arraybackslash}p{(\columnwidth - 6\tabcolsep) * \real{0.0833}}
  >{\raggedleft\arraybackslash}p{(\columnwidth - 6\tabcolsep) * \real{0.2500}}@{}}
\caption{\label{tbl-demographics}Demographics of the PPMI
dataset}\tabularnewline
\toprule\noalign{}
\endfirsthead
\endhead
\bottomrule\noalign{}
\endlastfoot
Group & Sex & N & Age {[}STD{]} \\
Control & F & 65 & 58.85 {[}11.95{]} \\
& M & 129 & 62.00 {[}10.74{]} \\
PD & F & 160 & 61.49 {[}9.96{]} \\
& M & 288 & 62.89 {[}9.71{]} \\
\end{longtable}

\hypertarget{sec-spa_norm}{%
\subsubsection{Spatial Normalization}\label{sec-spa_norm}}

Spatial normalization is frequently used in neuroimaging studies. It
eliminates differences in shape and size of brain, as well as local
inhomogeneities due to individual anatomic particularities. It is
particularly key in group analysis, where voxel-wise differences are
analysed and quantified (F. J. Martinez-Murcia, Górriz, and Ramírez
2016).

In this procedure, individual images are mapped from their individual
subject space (image space) to a common reference space, usually stated
using a template. The mapping involves the minimization of a cost
function that quantifies the differences between the individual image
space and the template. The most frequent template is the Montreal
Neurological Institute (MNI), set by the International Consortium for
Brain Mapping (ICBM) as its standard template, currently in its version
ICBM152 (Mazziotta et al. 2001), an average of 152 normal MRI scans in a
common space using a nine-parameter linear transformation.

There exist a wide range of global and local transformations that could
be categorized in linear transformations (of which the affine transform
is the most complex) and elastic transformations and diffeomorphisms.
The affine transform converts the old coordinates \((x,y,z)\) to the new
common coordinate system \((x',y',z')\) using 12 parameters for
translation, rotation, scale, squeeze, shear and others:
\begin{equation}\protect\hypertarget{eq-affine}{}{
\left[\begin{matrix}
x'\\y'\\z'\\1
\end{matrix}\right]
= \left[\begin{matrix}
a_{00} & a_{01} & a_{02} & a_{03}\\
a_{10} & a_{11} & a_{12} & a_{13}\\
a_{20} & a_{21} & a_{22} & a_{23}\\
0 & 0 & 0 & 1\\
\end{matrix}\right]
\left[\begin{matrix}
x\\y\\z\\1
\end{matrix}\right]}\label{eq-affine}\end{equation}

A particular case of affine transformation is the similarity
transformation, where only scale, translation and rotation are applied.
This is often used for motion correction and reorientation of brain
images with respect to a reference, and is frequently performed
automatically on many imaging equipment.

Affine transforms are applied globally to the whole image. In contrast,
modern solutions often refine the output by applying local deformations,
in what is known as diffeomorphic transformations, which feature the
estimation and application of a warp field. Some examples are SPM
(Friston et al. 2007) or Freesurfer (Reuter, Rosas, and Fischl 2010).

The DaTSCAN images from the PPMI dataset are roughly realigned. We will
refer to this as non-normalized (given that it is only a similarity
transformation that preserves shape). We further preprocessed the images
using the SPM12 New Normalize procedure with default parameters, which
applied affine and local deformations to achieve the best warping of the
images and a custom DaTSCAN template defined in Ref.~.

\hypertarget{sec-int_norm}{%
\subsection{Intensity Normalization}\label{sec-int_norm}}

Intensity normalization is a technique that changes globaly or locally
the intensity values of an image in order to ensure that the same
intensity levels correspond to similar physical measures. In nuclear
imaging, the use of intensity normalization is key in order to compare
brain activity or function between subjects. A similar intensity should
indicate a similar drug uptake and therefore, differences in these
values may be due to different pathologies (F. J. Martínez-Murcia et al.
2012; Fermín Segovia et al. 2012).

Intensity normalization in neuroimaging usually follows the expression:
\[\hat{\mathbf{I}}_i = \mathbf{I}_i/I_{n,i}\] where \(\mathbf{I}_i\) is
the image of the \(i^{th}\) subject in the dataset,
\(\hat{\mathbf{I}}_i\) is the normalized image, and \(I_{n}\) is an
intensity normalization value that is computed independently for each
subject.

Two normalization strategies will be used in this paper:

\begin{itemize}
\item
  In the \textbf{Normalization to the maximum}, \(I_n\) is computed as
  the average of the top 3\% intensities in an image. By averaging this
  top 3\% we ensure that the maximum intensity is not an outlier. The
  resulting image's intensities will be in the range \((0,1)\).
\item
  For its part, the \textbf{Integral Normalization} (I. A. a. Illán et
  al. 2012) sets \(I_n\) to the average of all values in a certain
  volume the image, in an approximation of the integral. In Parkinson,
  this is often set to the average of the brain without the specific
  areas: the striatum; although the influence of these areas is often
  small, and it can be approximated by the mean of the whole image. This
  is the procedure followed in this paper.
\end{itemize}

\hypertarget{convolutional-neural-networks}{%
\subsection{Convolutional Neural
Networks}\label{convolutional-neural-networks}}

Convolutional Neural Networks (CNNs) are becoming increasignly important
in the Machine Learning community (Krizhevsky, Sutskever, and Hinton
2012; Ciresan et al. 2011; Ortiz et al. 2016; Fermín Segovia et al.
2016; Sabour, Frosst, and Hinton 2017), especially within the artificial
vision and image analysis fields. Since 2012, when an ensemble of CNNs
(Krizhevsky, Sutskever, and Hinton 2012) achieved lowest error on the
ImageNet classification benchmark (Schmidhuber 2015), CNNs prevail over
any other pattern recognition algorithm in the literature in image
classification.

CNNs are bioinspired by the convolutional response of neurons, and
combine feature extraction and classification in one single
architecture. The combination of different convolution and pooling steps
is able to recognize different patterns, from low-level features to
higher abstractions, depending on the net depth. The set of fully
connected layers, similar to a perceptron, placed after convolutional
layers, performs the real machine learning. Many architectures combining
these and other types of layers can be found throughout the literature
(Krizhevsky, Sutskever, and Hinton 2012; Ciresan et al. 2011; Abadi et
al. 2015; Payan and Montana 2015; Ortiz et al. 2016; Fermín Segovia et
al. 2016; Sabour, Frosst, and Hinton 2017).

One key property that defines CNNs is parameter sharing. All neurons in
any convolutional layer share the same weights, saving memory and easing
computation of the convolutions. Other properties shared by CNNs are
local connectivity of the hidden units and the use of pooling to
introduce position invariance, although this latter feature is currently
being challenged by newer approaches (Springenberg et al. 2014; Sabour,
Frosst, and Hinton 2017)

\hypertarget{convolutional-layer}{%
\subsubsection{Convolutional Layer}\label{convolutional-layer}}

The convolutional layer is the main component of CNNs. Mathematically,
the operation at the convolutional layer takes a tensor
\(\mathbf{V}_{i-1}\) containing the activation map of the previous layer
\((i-1)\)-th (for \(i=1\) it is the input volume). The \(i^{th}\) layer
learns a set of \(K\) filters \(\mathbf{W}_{i}\) with a bias term
\(\mathbf{b}_i\), a vector of length \(K\). The mathematical operation
performed at the convolutional layer is:
\[\mathbf{V}_i = f_a \left(\mathbf{W}_{i}*\mathbf{V}_{i-1} + \mathbf{b}_i\right)\]
where \(f_a(*)\) is the activation function (see
Sec.~Section~\ref{sec-activations}).

For a three-dimensional environment, \(\mathbf{V}_{i-1}\) is of size
\(H\times W \times D \times C\) (height, width, depth and number of
channels), and \(\mathbf{W}_{i}\) is of size
\(P \times Q \times R \times S \times K\), with \(K\) the number of
filters. The \(k^{th}\) convolution term for the \(k^{th}\) filter is:

\begin{equation}\protect\hypertarget{eq-3dconv}{}{\begin{gathered}
\mathbf{W}_{ik}*\mathbf{V}_{i-1} =\\
  \sum_{u=0}^{P-1}\sum_{v=0}^{Q-1}\sum_{w=0}^{R-1} \big[\mathbf{W}_{ik}(P-u, Q-v, R-w)\cdot\\
  \mathbf{V}_{i-1}(x+u,y+v,z+w)\big]\end{gathered}}\label{eq-3dconv}\end{equation}

After convolution, the activations of the \(K\) filters in layer \(i\)
are stacked and passed to layer \((i+1)\).

There are a number of hyperparameters that must be set a priori, among
them the filter size (usually a cube of \(P=Q=R\)), which in the
literature take frequently values (LeCun et al. 1998; Krizhevsky,
Sutskever, and Hinton 2012) of 3, 5 or 7. Other parameters are the
number of filters \(K\), stride and zero-padding. The number of filters
varies a lot in the literature, frequently taking numbers that are a
power of 2. The higher this number is, the more patterns (simple in
lower layers, more complex in higher layers) our CNN is able to learn.
This number is usually high in two-dimensional CNNs (being 48, 96 or 128
frequent values for the first layer) (LeCun et al. 1998; Krizhevsky,
Sutskever, and Hinton 2012), and increasing in higher-order layers after
dropout. In three-dimensional layers, however, this number is smaller,
due to hardware requirements.

Stride, for its part, controls the step at which the convolution is
computer, namely how much overlapping there is between convolutions.
That defines the so called `receptive field' of a neuron, which is the
part of the image to which a neuron is connected. A stride of 1 means
that the convolution is performed at each voxel of the input. Higher
strides mean less overlapping between receptive fields and smaller
volumes in the output.

Finally, zero-padding provides control of the output volume by padding
the input volume with zeros. That way, we can let the output volume be
the same size as the input volume, which is desirable in some cases.

\hypertarget{sec-activations}{%
\subsubsection{Activations}\label{sec-activations}}

The activation is a function used to compute the output of a layer,
common to all types of ANNs. It is applied after a single value \(z\)
-weighted sum of inputs or the sum of all convolutions- is computed in
that layer, and is often consider to `fire' the neuron signal. Many
activation functions exist, among them the first function to be used in
ANNs: the hyperbolic tangent or \(\tanh\). In this work we use three
different activation functions.

First, in the convolutional layers we have chosen the Rectified Linear
Unit (ReLU), a non-saturating activation function. \[\label{eq:relu}
f_{RELU}(z)=\max(0,z)\] It has gained a lot of popularity thanks to its
properties, especially for making the training procedure much faster
than other approaches, since its derivative has a smaller computational
cost. It has been proven that this reduction in time comes without loss
of generalization ability (Krizhevsky, Sutskever, and Hinton 2012),
which makes it optimum for our purpose.

Second, we also wanted to assess the novel Self-Normalized Neural
Networks (SNNs) (Klambauer et al. 2017) in this context. SNNs are
designed in order to avoid vanishing and exploding gradients during the
training iterations, introducing the Scaled Exponential Linear Unit
(SELU): \[\label{eq:selu}
    f_{SELU}(z) = \lambda \begin{cases}
                z & \text{if } z\geq 0\\
                \alpha e^z-\alpha & \text{if } z<0\\
    \end{cases}\] This SELU function, used with the default parameters
calculated in Ref~ for the typical normalization \((\mu, \nu)=(0,1)\)
used in batch normalization, which correspond to \(\alpha\approx1.6733\)
and \(\lambda\approx 1.0507\). This function introduces self-normalizing
properties at the output of each layer which in the case of Feed Forward
Networks has proven to propagate zero-mean and unit-variance
gaussian-distributed outputs through the network even in noisy
environments.

The last activation function to be used is the well-established softmax,
used at the output layer. The softmax activation function for the
\(j^{th}\) neuron (devoted to the \(j^{th}\) class in the problem) in
the output layer follows the expression:
\[\sigma(\mathbf{z})_j = \frac{e^{z_j}}{\sum_{k=1}^K e^{z_k}}  \quad \text{for} j = 1, \dots K\]
where \(z_j=\mathbf{x}^T\mathbf{w}_j\), being \(\mathbf{x}\) the outputs
of the previous layer, and \(\mathbf{w}_j\) the weights of the
connections of the \(j^{th}\) neuron.

\hypertarget{max-pooling}{%
\subsubsection{Max Pooling}\label{max-pooling}}

A key operation in CNNs is pooling, which provides a sort of positional
invariance of the patterns that fire up neuron activations by reducing
the input space of subsequent layers while keeping the receptive field
of the filters. The most common type of pooling is Max Pooling
(MaxPool), which keeps the maximum value over a \(M\times M\times M\)
block of the activation layer in the case of 3D layers. Furthermore,
MaxPool also prevents the following layers from processing non-maximum
values, therefore reducing computational load.

Some works are starting to eliminate MaxPooling layers from CNN setups
(Springenberg et al. 2014), especially in recent capsule networks, where
they are being replaced by dynamic routing (Sabour, Frosst, and Hinton
2017). This provides the network with positional equivariance, which, in
contrast to the Pooling invariance, preserves the relative positions of
the different patterns detected at lower level layers. However, we have
opted to replicate well known models such as LENET and ALEXNET in
three-dimensional environments, and therefore, we keep the MaxPool
layers present in these models, with different block sizes.

\hypertarget{dense-layers}{%
\subsubsection{Dense Layers}\label{dense-layers}}

Fully connected layers, also known as dense layers, are the components
of typical ANN examples such as the multilayer perceptron (MLP). In
dense layers all neurons are connected to all outputs from the previous
layer. In CNNs, several dense layers are usually placed after a set of
convolution and pooling layers. They can be considered the high-level
reasoning part of the CNN. In these cases, the output of the \(j^{th}\)
neuron will be computed by: \[z_j=f(\mathbf{x}^T\mathbf{w}_j)\] where
\(f()\) is any activation function. In CNNs is frequent to have several
dense layers with RELU activation and finally, the last layer uses the
softmax activation function.

\hypertarget{sec-dropout}{%
\subsubsection{Dropout}\label{sec-dropout}}

Dropout is a method intended to reduce overfitting in fully connected
layers (Schmidhuber 2015). With this method, only a percentage \(1-p\)
of the neurons of one layer are active in one training iteration. In the
next iteration, the inactive neurons will recover their past weight
matrix, and the procedure is repeated. At testing time, all neurons are
active, and therefore the output of the neurons is weighted by a factor
of \(p\), approximating the use of all possible \(2^n\) networks.

A particular case of dropout is AlphaDropout (Klambauer et al. 2017),
devised originally for SNNs. It keeps the mean and variance of inputs to
their original values, therefore ensuring the self-normalizing
assumption even after dropout. Instead of turning off the neurons,
establishes their weight as the negative saturation value of the
function.

\hypertarget{sec-loss}{%
\subsubsection{Loss functions}\label{sec-loss}}

The CNN is trained with the objective of optimizing a loss function.
This is often chosen as a result of trial and error, and is hardly
reported in many CNN papers. In this work, we have tried two different
functions with different properties, in order to see their influence in
the convergence of the models.

The first strategy here is to maximize the cross-entropy (x-e), a
measure of similarity between the real output \(y_i\) and the predicted
output \(\hat{y}_i\) of the classifier for all \(N\) images in a
training batch. This is defined by:
\[\ell_{x-e}=\ -\frac1N\sum_{i=1}^N\ \bigg[y_i  \log \hat y_i \bigg]\]

A second approach, more frequent in regression but gaining ground in
classification, is logcosh (lc), or the \(\log(cosh(x))\). It is similar
to other functions, but twice differentiable everywhere, and
approximately equal to \((x^2)/2\) for small x and to
\(\text{abs}(x) - \log(2)\) for large x. It has the advantage of working
similarly to the mean squared error, but being not so strongly affected
by occasional wrong predictions:
\[\ell_{lc} = -\frac1N\sum_{i=1}^N\bigg[\log(\cosh(\hat y_i -y_i))\bigg]\]

\hypertarget{our-models}{%
\subsection{Our models}\label{our-models}}

\begin{figure}

{\centering \includegraphics{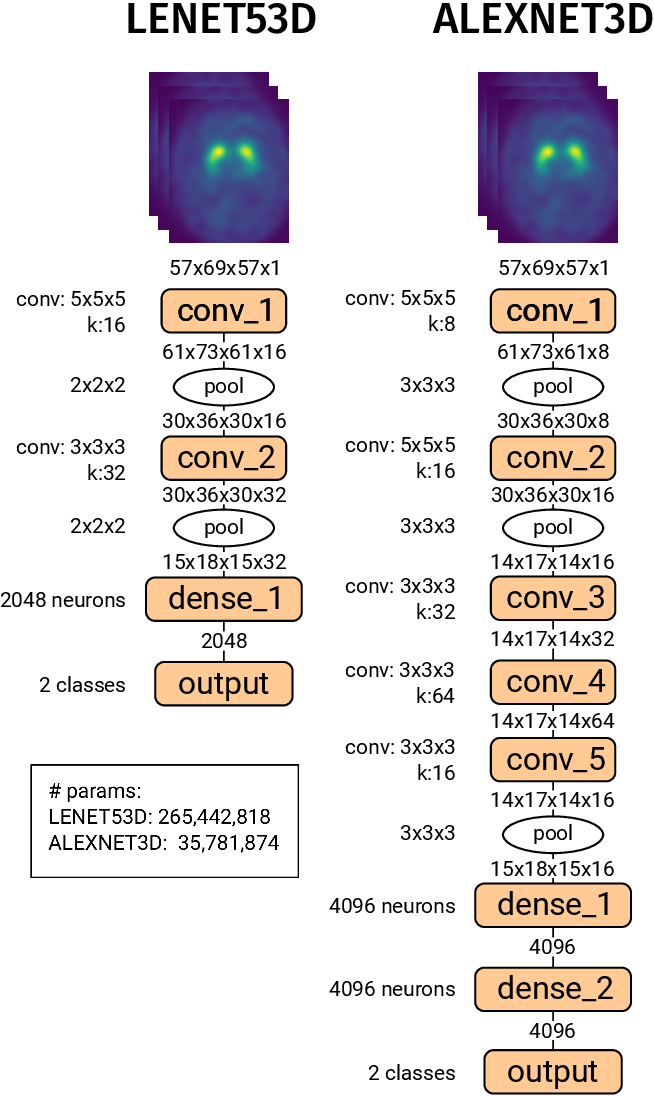}

}

\caption{\label{fig-schema}Outline of the LENET53D and ALEXNET3D models
used in this article, including the size of convolutions and pooling,
number of filters, size of tensors and total number of parameters.}

\end{figure}

We have tested two different CNN models, based on a three-dimensional
implementation of well-established architectures. We assume the
three-dimensional convolutions stated in Equation~\ref{eq-3dconv}, and a
three-dimensional max pooling. The number of filters and number of
neurons in dense layers is smaller than in their 2D counterparts, mainly
due to the memory restrictions and the fact that this is a binary
classification problem:

\begin{itemize}
\item
  LENET53D is a 3D implementation of the LeNet-5 (LeCun et al. 1998) as
  in the original paper architecture. This architecture is composed of 5
  layers, combining 2 convolutional, 2 max-pooling and 1 fully connected
  layer as in the structure found in Figure~\ref{fig-schema}.
\item
  ALEXNET3D is a 3D implementation of the AlexNet (Krizhevsky,
  Sutskever, and Hinton 2012), which uses five convolutional layers, two
  fully connected layers and three max-pooling layers, following the
  architecture found at Figure~\ref{fig-schema}.
\end{itemize}

These two models have been also implemented with the SELU activation
function. These two models, LENET53D-SELU and ALEXNET3D-SELU, have the
same number of parameters that their SELU counterparts, and only differ
in the activation function, the initialization of the weights at each
layer, and the dropout technique suggested in Ref~.

\hypertarget{evaluation}{%
\subsubsection{Evaluation}\label{evaluation}}

Multiple evaluation parameters have been used in the results section,
considering a binary classification problem (PD vs.~controls): accuracy,
sensitivity, specificity, balanced accuracy and F1-score -all derived
from the confusion matrix-, the Receiver Operating Characteristic (ROC)
curve, and the Area Under the ROC Curve (AUC). The confusion matrix
contains the number of True Positives (TP), True Negatives (TN), False
Positives (FP) and False Negatives (FN) of the prediction of the trained
model, from which we can compute the parameters: \[\begin{aligned}
\text{acc} & = \frac{TP + TN}{TP + TN + FP + FN}\\
\text{sens} & = \frac{TP}{TP + FN}\\
\text{spec} & = \frac{TN}{TN + FP}\\
\text{F1-score} & = \frac{2TP}{2TP+FN+FP}\\
\text{bal. acc} & =\frac{\text{sens}+\text{spec}}{2}\end{aligned}\]

These parameters have been computed within a cross-validation scheme,
namely a 10-fold stratified cross-validation, where the distribution of
classes within each fold is similar to the distribution of classes in
the whole dataset (Kohavi and John 1995). We have trained the different
models over 60 epochs with a batch size of 64, and using class weights
equal to the proportion of samples in each class (due to our imbalanced
dataset).

The ROC curve depicts the sensitivity over one minus the specificity at
different thresholds of a classifier. This curve can be computed by
ranking the softmax values at the output layer for each class, and
depicting each point depending on whether it is a TP or FP (Hanley and
McNeil 1982). It yields a visual comparison of different methods,
frequently used in many works (F. J. Martínez-Murcia et al. 2012; F.
Segovia et al. 2012; I. A. a. Illán et al. 2012). We can integrate each
curve to obtain the AUC, another estimate of the overall performance,
which is reported for each curve. Note that we will use a name
convention of {[}no\(|\)int\(|\)max{]}\_{[}u\(|\)w{]} to abbreviate
legends in the figures of the result section, where `no', `int', and
`max' represent the types of intensity normalization and `w' and `u'
state if we apply or not spatial normalization respectively.

For assessing the most relevant regions, we use saliency maps (Simonyan,
Vedaldi, and Zisserman 2013) which, in brief, highlight the salient
image regions that contribute the most to the assigned class. This is
done by computing the gradient of the output category with respect to a
sample input image as in:
\[\frac{\partial \text{output}}{\partial I_\text{input}}\] This
quantifies the changes in the output score with respect to a small
change in the input. This eventually leads to a map the same size of the
original images where the most relevant voxels are highlighted. To
generate the maps we have used the keras-vis (Kotikalapudi and
contributors 2017) python toolbox.

\hypertarget{sec-results}{%
\section{Results}\label{sec-results}}

Each proposed model has been trained and tested via cross-validation on
a 642-subject dataset containing DaTSCAN images from PD and normal
control. For more details on the demographics, please check
Table~\ref{tbl-demographics}. The particular results for each model are
detailed below.

\hypertarget{results-for-the-lenet53d-model}{%
\subsection{Results for the LENET53D
model}\label{results-for-the-lenet53d-model}}

We use a LENET model analogous to the first five-layer LENET5 proposed
in Ref.~, although using 3D convolutions instead of 2D. The performance
of this model, depending on the preprocessing, and loss function applied
can be checked at Table~\ref{tbl-lenet_perf_all}.

\hypertarget{tbl-lenet_perf_all}{}
\begin{longtable}[]{@{}
  >{\raggedright\arraybackslash}p{(\columnwidth - 14\tabcolsep) * \real{0.1111}}
  >{\raggedright\arraybackslash}p{(\columnwidth - 14\tabcolsep) * \real{0.1111}}
  >{\raggedright\arraybackslash}p{(\columnwidth - 14\tabcolsep) * \real{0.0972}}
  >{\centering\arraybackslash}p{(\columnwidth - 14\tabcolsep) * \real{0.1111}}
  >{\centering\arraybackslash}p{(\columnwidth - 14\tabcolsep) * \real{0.1111}}
  >{\centering\arraybackslash}p{(\columnwidth - 14\tabcolsep) * \real{0.1111}}
  >{\centering\arraybackslash}p{(\columnwidth - 14\tabcolsep) * \real{0.1111}}
  >{\centering\arraybackslash}p{(\columnwidth - 14\tabcolsep) * \real{0.1111}}@{}}
\caption{\label{tbl-lenet_perf_all}Performance values and standard
deviation of the different preprocessing pipelines in the LENET53D
model.}\tabularnewline
\toprule\noalign{}
\endfirsthead
\endhead
\bottomrule\noalign{}
\endlastfoot
S -norm & I- norm. & loss & acc {[} std{]} & sens. {[} std{]} & spec.
{[} std{]} & f1 & bal. acc \\
yes & no & x-e & 0.500 {[}0. 198{]} & 0.500 {[}0. 500{]} & 0.500 {[}0.
500{]} & 0.500 & 0.500 \\
yes & no & lc & 0.461 {[}0. 194{]} & 0.400 {[}0. 490{]} & 0.600 {[}0.
500{]} & 0.444 & 0.500 \\
yes & max & x-e & 0.760 {[}0. 072{]} & 0.922 {[}0. 090{]} & 0.389 {[}0.
318{]} & 0.728 & 0.656 \\
yes & max & lc & 0.791 {[}0. 087{]} &
\begin{minipage}[t]{\linewidth}\centering
\textbf{ 0.929\\
{[}0.08 1{]}}\strut
\end{minipage} & 0.474 {[}0. 257{]} & 0.757 & 0.702 \\
yes & int & x-e & 0.872 {[}0. 109{]} & 0.891 {[}0. 139{]} & 0.827 {[}0.
228{]} & 0.864 & 0.859 \\
yes & int & lc & \begin{minipage}[t]{\linewidth}\centering
\textbf{ 0.922\\
{[}0.03 7{]}}\strut
\end{minipage} & 0.918 {[}0. 054{]} &
\begin{minipage}[t]{\linewidth}\centering
\textbf{ 0.932\\
{[}0.08 6{]}}\strut
\end{minipage} & \textbf{0. 924} & \textbf{0. 925} \\
no & no & x-e & 0.464 {[}0. 199{]} & 0.369 {[}0. 460{]} & 0.685 {[}0.
482{]} & 0.438 & 0.527 \\
no & no & lc & 0.494 {[}0. 193{]} & 0.487 {[}0. 488{]} & 0.510 {[}0.
490{]} & 0.492 & 0.498 \\
no & max & x-e & 0.494 {[}0. 192{]} & 0.449 {[}0. 458{]} & 0.597 {[}0.
456{]} & 0.485 & 0.523 \\
no & max & lc & 0.866 {[}0. 205{]} & 0.860 {[}0. 288{]} & 0.879 {[}0.
292{]} & 0.868 & 0.869 \\
no & int & x-e & 0.575 {[}0. 191{]} & 0.561 {[}0. 351{]} & 0.604 {[}0.
345{]} & 0.574 & 0.583 \\
no & int & lc & \begin{minipage}[t]{\linewidth}\centering
\textbf{ 0.949\\
{[}0.02 5{]}}\strut
\end{minipage} & \begin{minipage}[t]{\linewidth}\centering
\textbf{ 0.940\\
{[}0.04 6{]}}\strut
\end{minipage} & \begin{minipage}[t]{\linewidth}\centering
\textbf{ 0.969\\
{[}0.05 1{]}}\strut
\end{minipage} & \textbf{0. 954} & \textbf{0. 954} \\
\end{longtable}

\begin{figure}

{\centering \includegraphics{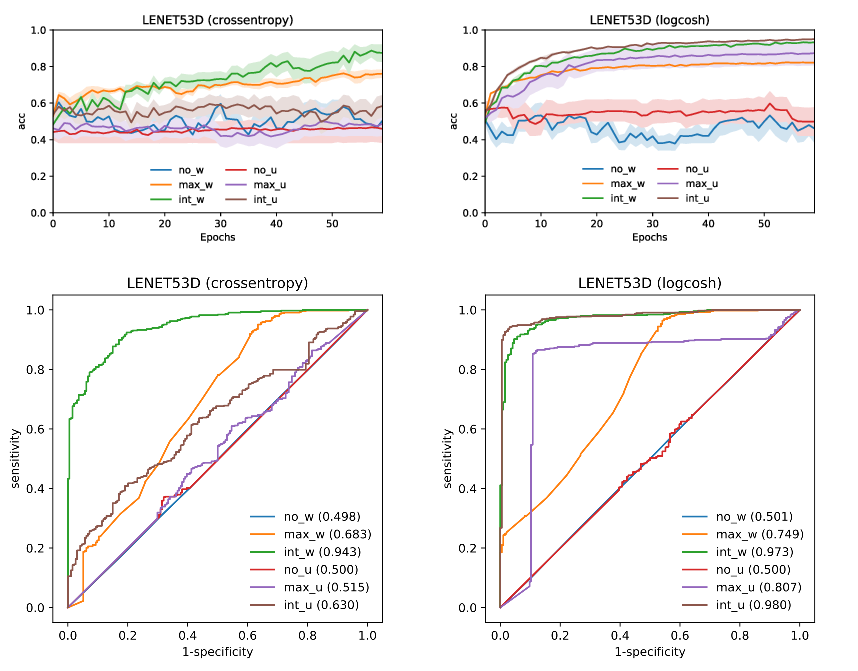}

}

\caption{\label{fig-lenet5-perf}Performance of the different
normalization methods and loss functions in the LENET53D architecture.
Upper part: evolution (mean and standard deviation) of the training
accuracy in different epochs; lower part: ROC curves at the operation
point.}

\end{figure}

The behaviour of the LENET53D model is quite complex. When there is
spatial normalization, the influence of the loss function is minimal,
although the logcosh (lc) often performs better than the cross-entropy
(x-e). The behaviour is, however, very different when there is no
spatial normalization applied. In this case, the logcosh achieves
significantly better performance than the models trained with
cross-entropy, in any intensity normalization case. This behaviour is
even more clear in @\#fig-lenet5-perf, where the convergence of the
system using this function is clearly faster than when using
cross-entropy, for any combination of spatial and intensity
normalization. This even helps the model to perform similarly regardless
of the intensity normalization.

Regarding the data, intensity normalization may be key for the model to
correctly account for PKS-related patterns. Of all intensity
normalization strategies, the integral normalization always perform
better than the normalization to the maximum, even when there is no
spatial normalization applied.

\hypertarget{results-for-the-lenet53d-selu-model}{%
\subsection{Results for the LENET53D-SELU
model}\label{results-for-the-lenet53d-selu-model}}

The LENET53D-SELU is identical to the LENET53D, but using SELU as
activation function, a different random initialization and the custom
dropout defined at Section~\ref{sec-activations} and
Section~\ref{sec-dropout}.

\hypertarget{tbl-lenet_selu_perf_all}{}
\begin{longtable}[]{@{}
  >{\raggedright\arraybackslash}p{(\columnwidth - 14\tabcolsep) * \real{0.1111}}
  >{\raggedright\arraybackslash}p{(\columnwidth - 14\tabcolsep) * \real{0.1111}}
  >{\raggedright\arraybackslash}p{(\columnwidth - 14\tabcolsep) * \real{0.0972}}
  >{\centering\arraybackslash}p{(\columnwidth - 14\tabcolsep) * \real{0.1111}}
  >{\centering\arraybackslash}p{(\columnwidth - 14\tabcolsep) * \real{0.1111}}
  >{\centering\arraybackslash}p{(\columnwidth - 14\tabcolsep) * \real{0.1111}}
  >{\centering\arraybackslash}p{(\columnwidth - 14\tabcolsep) * \real{0.1111}}
  >{\centering\arraybackslash}p{(\columnwidth - 14\tabcolsep) * \real{0.1111}}@{}}
\caption{\label{tbl-lenet_selu_perf_all}Performance values and standard
deviation of the different preprocessing pipelines in the LENET53D-SELU
model.}\tabularnewline
\toprule\noalign{}
\endfirsthead
\endhead
\bottomrule\noalign{}
\endlastfoot
S -norm & I- norm. & loss & acc {[} std{]} & sens. {[} std{]} & spec.
{[} std{]} & f1 & bal. acc \\
yes & no & x-e & 0.302 {[}0. 005{]} & 0.000 {[}0. 000{]} & 1.000 {[}0.
500{]} & nan & 0.500 \\
yes & no & lc & 0.302 {[}0. 005{]} & 0.000 {[}0. 000{]} & 1.000 {[}0.
500{]} & nan & 0.500 \\
yes & max & x-e & 0.511 {[}0. 146{]} & 0.516 {[}0. 318{]} & 0.499 {[}0.
303{]} & 0.511 & 0.507 \\
yes & max & lc & \begin{minipage}[t]{\linewidth}\centering
\textbf{ 0.617\\
{[}0.06 7{]}}\strut
\end{minipage} & \begin{minipage}[t]{\linewidth}\centering
\textbf{ 0.665\\
{[}0.08 8{]}}\strut
\end{minipage} & 0.506 {[}0. 143{]} & \textbf{0. 616} & 0.585 \\
yes & int & x-e & 0.460 {[}0. 194{]} & 0.400 {[}0. 490{]} & 0.600 {[}0.
500{]} & 0.444 & 0.500 \\
yes & int & lc & 0.556 {[}0. 184{]} & 0.447 {[}0. 345{]} &
\begin{minipage}[t]{\linewidth}\centering
\textbf{ 0.808\\
{[}0.37 6{]}}\strut
\end{minipage} & 0.545 & \textbf{0. 627} \\
no & no & x-e & 0.540 {[}0. 194{]} & 0.600 {[}0. 490{]} & 0.400 {[}0.
500{]} & 0.545 & 0.500 \\
no & no & lc & 0.542 {[}0. 193{]} & 0.600 {[}0. 490{]} & 0.400 {[}0.
500{]} & 0.545 & 0.500 \\
no & max & x-e & \begin{minipage}[t]{\linewidth}\centering
\textbf{ 0.618\\
{[}0.15 8{]}}\strut
\end{minipage} & \begin{minipage}[t]{\linewidth}\centering
\textbf{ 0.800\\
{[}0.40 0{]}}\strut
\end{minipage} & 0.200 {[}0. 500{]} & \textbf{0. 615} & 0.500 \\
no & max & lc & 0.581 {[}0. 195{]} & 0.593 {[}0. 438{]} & 0.553 {[}0.
451{]} & 0.581 & 0.573 \\
no & int & x-e & 0.421 {[}0. 181{]} & 0.300 {[}0. 458{]} & 0.700 {[}0.
500{]} & 0.375 & 0.500 \\
no & int & lc & 0.609 {[}0. 252{]} & 0.587 {[}0. 472{]} &
\begin{minipage}[t]{\linewidth}\centering
\textbf{ 0.663\\
{[}0.42 8{]}}\strut
\end{minipage} & 0.610 & \textbf{0. 625} \\
\end{longtable}

\begin{figure}

{\centering \includegraphics{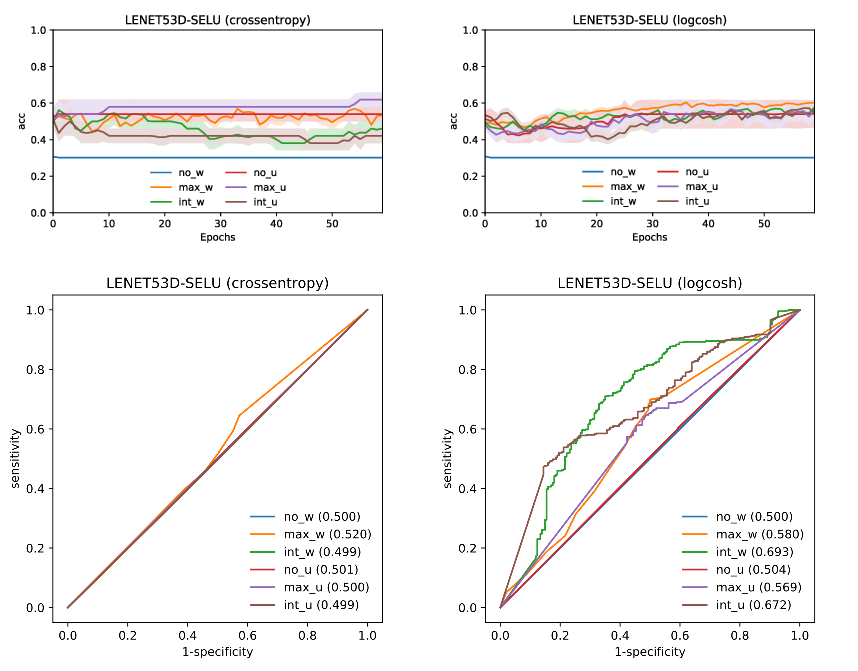}

}

\caption{\label{fig-lenet5-selu-perf}Performance of the different
normalization methods and loss functions in the LENET53D-SELU
architecture. Upper part: evolution (mean and standard deviation) of the
training accuracy in different epochs; lower part: ROC curves at the
operation point.}

\end{figure}

The results are much poorer than in its regular ReLU counterpart. There
is no sign of convergence when using the cross-entropy loss function, as
seen in Figure~\ref{fig-lenet5-selu-perf}, nor any difference with a
random classifier (see ROC curve). By looking at the sensitivities and
specificities of Table~\ref{tbl-lenet_selu_perf_all}, it is evident that
the classifier is biased, yielding sensitivities multiples of 0.1, which
is probably taking all samples as any of the classes at random in each
cross-validation iteration.

The logcosh function, although still biased, makes for a better
predictor, particularly in the case of integral normalization and
spatially normalized images. However, the convergence, if any, is very
slow, and after 60 training epochs there is no evidence that the loss is
going to decrease. Still, compared to the crossentropy loss function,
the logcosh makes the system slightly evolve, which may be an indication
of possible learning.

\hypertarget{results-for-the-alexnet3d-model}{%
\subsection{Results for the ALEXNET3D
model}\label{results-for-the-alexnet3d-model}}

Table~\ref{tbl-alex_perf_all_ce} shows the cross-validated performance
values obtained for the database under each preprocessing scheme. Under
this model, intensity normalization seems key to a good performance. The
best performance is obtained with the images using integral intensity
normalization, regardless of the spatial normalization. In the case of
un-normalized images, the accuracy is almost 94\%, with the highest
specificity rate -which is key in this unbalanced dataset-. However,
when using registered images, the performance does not drop
significantly, and in the case of the images normalized to the maximum,
the performance even increases.

\hypertarget{tbl-alex_perf_all_ce}{}
\begin{longtable}[]{@{}
  >{\raggedright\arraybackslash}p{(\columnwidth - 14\tabcolsep) * \real{0.1111}}
  >{\raggedright\arraybackslash}p{(\columnwidth - 14\tabcolsep) * \real{0.1111}}
  >{\raggedright\arraybackslash}p{(\columnwidth - 14\tabcolsep) * \real{0.0972}}
  >{\centering\arraybackslash}p{(\columnwidth - 14\tabcolsep) * \real{0.1111}}
  >{\centering\arraybackslash}p{(\columnwidth - 14\tabcolsep) * \real{0.1111}}
  >{\centering\arraybackslash}p{(\columnwidth - 14\tabcolsep) * \real{0.1111}}
  >{\centering\arraybackslash}p{(\columnwidth - 14\tabcolsep) * \real{0.1111}}
  >{\centering\arraybackslash}p{(\columnwidth - 14\tabcolsep) * \real{0.1111}}@{}}
\caption{\label{tbl-alex_perf_all_ce}Performance values and standard
deviation of the different preprocessing pipelines in the ALEXNET3D
model.}\tabularnewline
\toprule\noalign{}
\endfirsthead
\endhead
\bottomrule\noalign{}
\endlastfoot
S -norm & I- norm. & loss & acc {[} std{]} & sens. {[} std{]} & spec.
{[} std{]} & f1 & bal. acc \\
yes & no & x-e & 0.498 {[}0. 198{]} & 0.500 {[}0. 500{]} & 0.500 {[}0.
500{]} & 0.500 & 0.500 \\
yes & no & lc & 0.500 {[}0. 198{]} & 0.500 {[}0. 500{]} & 0.500 {[}0.
500{]} & 0.500 & 0.500 \\
yes & max & x-e & 0.793 {[}0. 064{]} & 0.922 {[}0. 053{]} & 0.496 {[}0.
239{]} & 0.760 & 0.709 \\
yes & max & lc & 0.788 {[}0. 042{]} &
\begin{minipage}[t]{\linewidth}\centering
\textbf{ 0.967\\
{[}0.02 9{]}}\strut
\end{minipage} & 0.378 {[}0. 320{]} & 0.747 & 0.672 \\
yes & int & x-e & 0.897 {[}0. 177{]} & 0.869 {[}0. 234{]} & 0.964 {[}0.
175{]} & 0.912 & 0.916 \\
yes & int & lc & \begin{minipage}[t]{\linewidth}\centering
\textbf{ 0.907\\
{[}0.16 4{]}}\strut
\end{minipage} & 0.880 {[}0. 208{]} &
\begin{minipage}[t]{\linewidth}\centering
\textbf{ 0.968\\
{[}0.16 0{]}}\strut
\end{minipage} & \textbf{0. 921} & \textbf{0. 924} \\
no & no & x-e & 0.556 {[}0. 174{]} & 0.649 {[}0. 423{]} & 0.341 {[}0.
447{]} & 0.562 & 0.495 \\
no & no & lc & 0.572 {[}0. 155{]} & 0.662 {[}0. 359{]} & 0.362 {[}0.
384{]} & 0.575 & 0.512 \\
no & max & x-e & 0.617 {[}0. 215{]} & 0.572 {[}0. 308{]} & 0.719 {[}0.
306{]} & 0.617 & 0.645 \\
no & max & lc & 0.601 {[}0. 235{]} & 0.598 {[}0. 428{]} & 0.608 {[}0.
425{]} & 0.601 & 0.603 \\
no & int & x-e & \begin{minipage}[t]{\linewidth}\centering
\textbf{ 0.941\\
{[}0.04 5{]}}\strut
\end{minipage} & \begin{minipage}[t]{\linewidth}\centering
\textbf{ 0.933\\
{[}0.05 8{]}}\strut
\end{minipage} & 0.958 {[}0. 079{]} & \textbf{0. 945} & \textbf{0.
946} \\
no & int & lc & 0.931 {[}0. 053{]} & 0.915 {[}0. 079{]} &
\begin{minipage}[t]{\linewidth}\centering
\textbf{ 0.969\\
{[}0.07 2{]}}\strut
\end{minipage} & 0.941 & 0.942 \\
\end{longtable}

\begin{figure}

{\centering \includegraphics{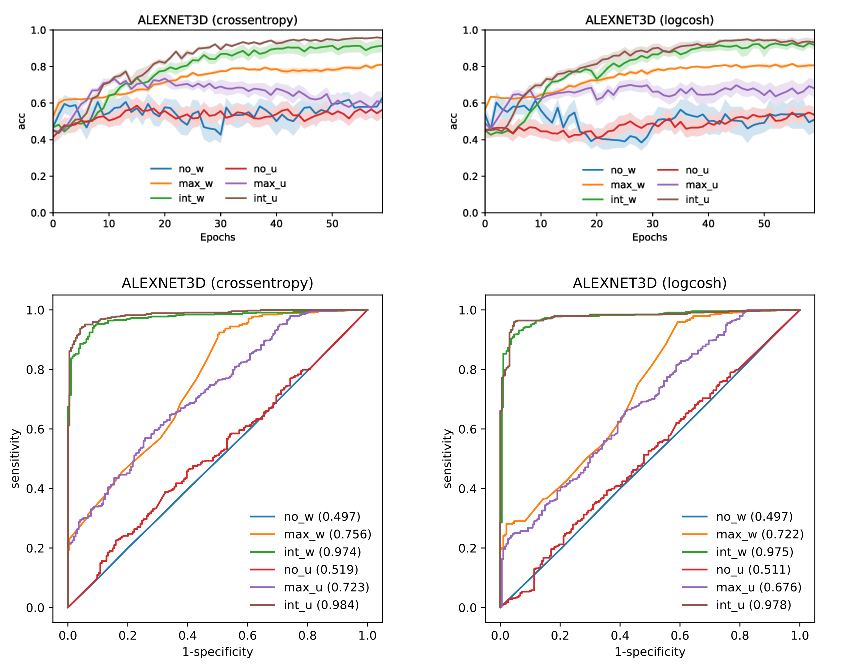}

}

\caption{\label{fig-alexnet3d-perf}Performance of the different
normalization methods and loss functions in the ALEXNET3D architecture.
Upper part: evolution (mean and standard deviation) of the training
accuracy in different epochs; lower part: ROC curves at the operation
point.}

\end{figure}

The loss function has little impact in this case, as it can be seen from
Table~\ref{tbl-alex_perf_all_ce} and Figure~\ref{fig-alexnet3d-perf}.
The convergence is similar both with cross-entropy and logcosh, and the
AUC is similarly high in any case that uses integral intensity
normalization, followed by normalization to the maximum, and values
approximately of a random classifier when not using intensity
normalization.

\hypertarget{results-for-the-alexnet3d-selu-model}{%
\subsection{Results for the ALEXNET3D-SELU
model}\label{results-for-the-alexnet3d-selu-model}}

The SELU counterpart of the ALEXNET3D drops the performance again, as it
can be seen in Table~\ref{tbl-alex_selu_perf_all_ce}. Its poorer
results, regardless of spatial or intensity normalization, can again be
attributed to the classifier randomly assigning the same class to all
samples in each cross-validation iteration.

\hypertarget{tbl-alex_selu_perf_all_ce}{}
\begin{longtable}[]{@{}
  >{\raggedright\arraybackslash}p{(\columnwidth - 14\tabcolsep) * \real{0.1111}}
  >{\raggedright\arraybackslash}p{(\columnwidth - 14\tabcolsep) * \real{0.1111}}
  >{\raggedright\arraybackslash}p{(\columnwidth - 14\tabcolsep) * \real{0.0972}}
  >{\centering\arraybackslash}p{(\columnwidth - 14\tabcolsep) * \real{0.1111}}
  >{\centering\arraybackslash}p{(\columnwidth - 14\tabcolsep) * \real{0.1111}}
  >{\centering\arraybackslash}p{(\columnwidth - 14\tabcolsep) * \real{0.1111}}
  >{\centering\arraybackslash}p{(\columnwidth - 14\tabcolsep) * \real{0.1111}}
  >{\centering\arraybackslash}p{(\columnwidth - 14\tabcolsep) * \real{0.1111}}@{}}
\caption{\label{tbl-alex_selu_perf_all_ce}Performance values and
standard deviation of the different preprocessing pipelines in the
ALEXNET3D-SELU model.}\tabularnewline
\toprule\noalign{}
\endfirsthead
\endhead
\bottomrule\noalign{}
\endlastfoot
S -norm & I- norm. & loss & acc {[} std{]} & sens. {[} std{]} & spec.
{[} std{]} & f1 & bal. acc \\
yes & no & x-e & 0.302 {[}0. 005{]} & 0.000 {[}0. 000{]} & 1.000 {[}0.
500{]} & nan & 0.500 \\
yes & no & lc & 0.302 {[}0. 005{]} & 0.000 {[}0. 000{]} & 1.000 {[}0.
500{]} & nan & 0.500 \\
yes & max & x-e & 0.414 {[}0. 172{]} & 0.276 {[}0. 426{]} & 0.735 {[}0.
479{]} & 0.358 & 0.505 \\
yes & max & lc & 0.618 {[}0. 068{]} & 0.667 {[}0. 091{]} & 0.506 {[}0.
144{]} & 0.617 & 0.587 \\
yes & int & x-e & 0.659 {[}0. 119{]} &
\begin{minipage}[t]{\linewidth}\centering
\textbf{ 0.900\\
{[}0.30 0{]}}\strut
\end{minipage} & 0.100 {[}0. 500{]} & 0.643 & 0.500 \\
yes & int & lc & \begin{minipage}[t]{\linewidth}\centering
\textbf{ 0.670\\
{[}0.29 7{]}}\strut
\end{minipage} & 0.564 {[}0. 453{]} &
\begin{minipage}[t]{\linewidth}\centering
\textbf{ 0.916\\
{[}0.37 7{]}}\strut
\end{minipage} & \textbf{0. 685} & \textbf{0. 740} \\
no & no & x-e & \begin{minipage}[t]{\linewidth}\centering
\textbf{ 0.581\\
{[}0.18 1{]}}\strut
\end{minipage} & \begin{minipage}[t]{\linewidth}\centering
\textbf{ 0.700\\
{[}0.45 8{]}}\strut
\end{minipage} & 0.300 {[}0. 500{]} & \textbf{0. 583} & 0.500 \\
no & no & lc & 0.537 {[}0. 194{]} & 0.600 {[}0. 490{]} & 0.400 {[}0.
500{]} & 0.545 & 0.500 \\
no & max & x-e & 0.540 {[}0. 194{]} & 0.600 {[}0. 490{]} & 0.400 {[}0.
500{]} & 0.545 & 0.500 \\
no & max & lc & 0.537 {[}0. 194{]} & 0.600 {[}0. 490{]} & 0.400 {[}0.
500{]} & 0.545 & 0.500 \\
no & int & x-e & 0.498 {[}0. 198{]} & 0.500 {[}0. 500{]} & 0.500 {[}0.
500{]} & 0.500 & 0.500 \\
no & int & lc & 0.505 {[}0. 257{]} & 0.351 {[}0. 444{]} &
\begin{minipage}[t]{\linewidth}\centering
\textbf{ 0.862\\
{[}0.44 8{]}}\strut
\end{minipage} & 0.472 & \textbf{0. 607} \\
\end{longtable}

\begin{figure}

{\centering \includegraphics{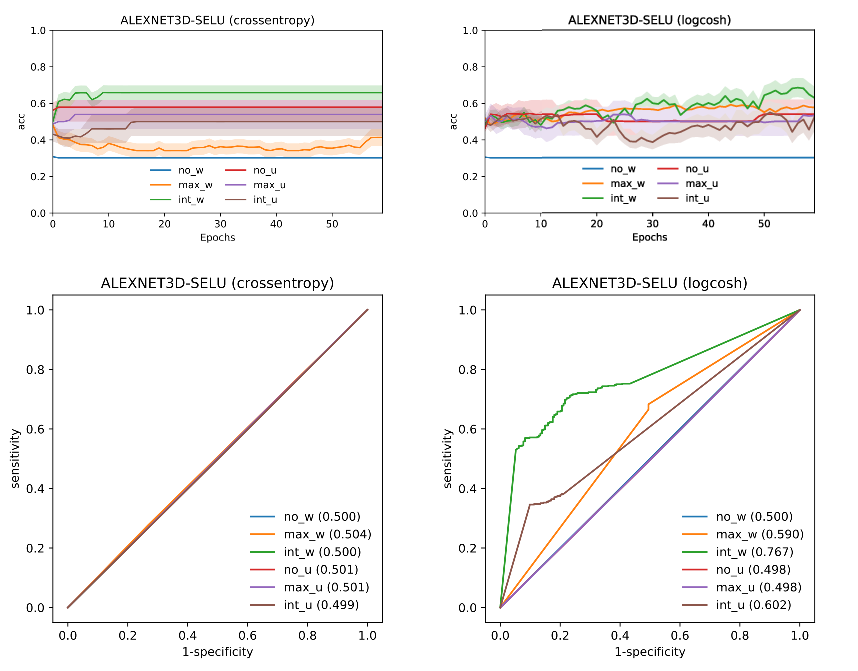}

}

\caption{\label{fig-alexnet3d-selu-perf}Performance of the different
normalization methods and loss functions in the ALEXNET3D-SELU
architecture. Upper part: evolution (mean and standard deviation) of the
training accuracy in different epochs; lower part: ROC curves at the
operation point.}

\end{figure}

The main difference can be seen when comparing loss functions, as
depicted in Figure~\ref{fig-alexnet3d-selu-perf}. While the
cross-entropy makes the system hardly evolve with the training
iterations, the logcosh seems to be increasing the performance, although
very slowly. After 60 iterations, the system achieves an AUC of 0.76
when using spatially and intensity normalized images, which is still an
improvement over the other preprocessing strategies.

\hypertarget{sec-discussion}{%
\section{Discussion}\label{sec-discussion}}

CNNs are gaining ground in the computer vision community, thanks in part
to their ability to recognize patterns with positional invariance. This
key feature could be exported to neuroimaging analysis, where spatial
normalization has been the norm in the latest 20 years, mainly due to
the prevalence of voxel-wise analyses (F. J. Martinez-Murcia, Górriz,
and Ramírez 2016).

MaxPooling is partially responsible for this positional invariance. It
allows the combination of different simple patterns, computed in the
first layers, to account for complex patterns found in latter layers. An
increasing number of voices are raising against MaxPool (Springenberg et
al. 2014; Sabour, Frosst, and Hinton 2017), pointing to this positional
invariance is not equivariance, and that a similar combination of
low-level features without a knowledge of their relative position could
lead to misclassification (Sabour, Frosst, and Hinton 2017). In this
context, capsule networks (Sabour, Frosst, and Hinton 2017) established
dynamic routing to account for these differences. However, these are
still useful, especially in environments such as neuroimaging, where the
images are very similar. In these images the systems find subtle
differences in function and/or anatomy, and large-scale artifacts such
as switching lobes are very unlikely to occur. Furthermore, MaxPool
could be of great help in images where smoothness is an inherent quality
(Boureau, Ponce, and LeCun 2010), by selecting larger features without
loss of generality.

We have tested two different models, with two variations in each,
involving the ReLU and SELU activation functions. Overall, the models
using SELU feature a poorer performance, similar to a classifier
randomly assigning a certain class to all inputs. Subtle increments in
performance can be appreciated with the logcosh loss function, in
contrast to cross-entropy. This differs from the original paper using
SELU (Klambauer et al. 2017), which used cross-entropy as learning
function. While the SELU function was proposed to overcome problems such
as vanishing and exploding gradients associated to ReLU in deep dense
networks, we could not prove any improvement in our deep CNN. Only the
ALEXNET3D-SELU model started to show some performance increment after 60
epochs of training, which by any means indicates a much slower
convergence than their ReLU counterparts.

\begin{figure}

{\centering \includegraphics{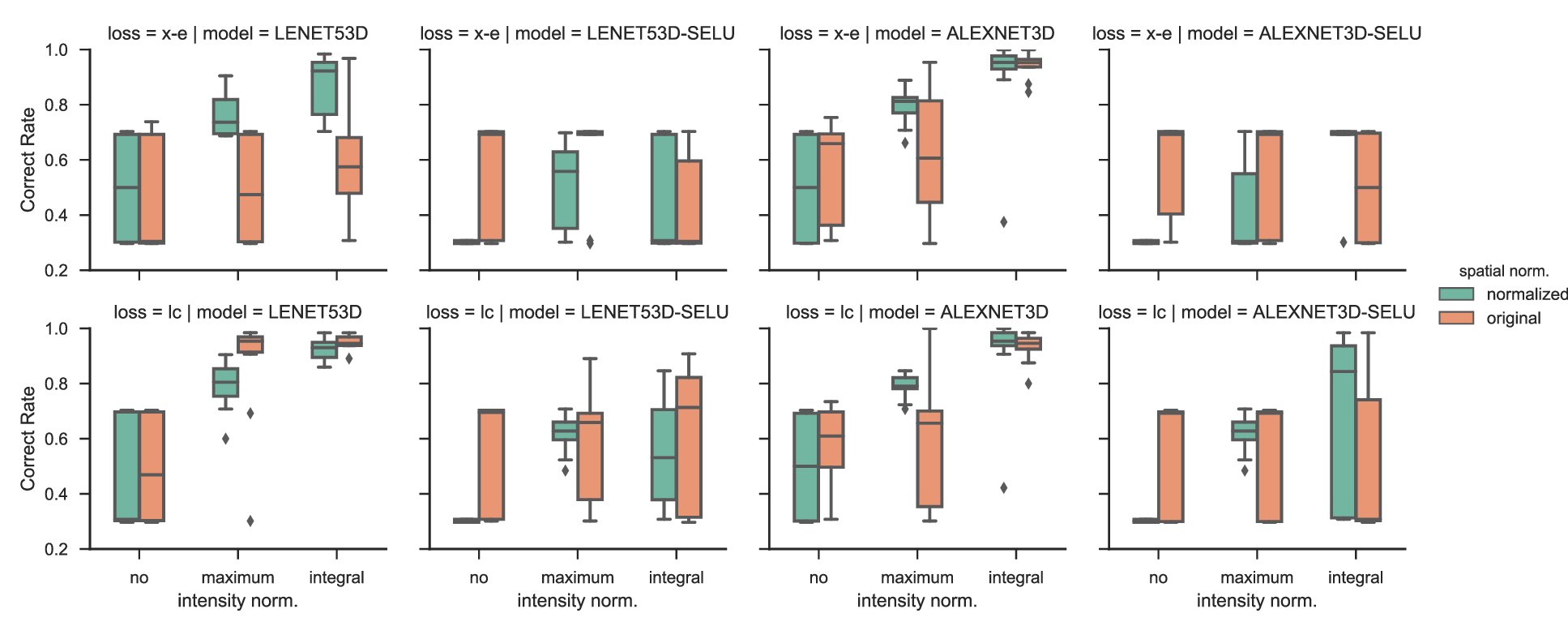}

}

\caption{\label{fig-boxplot}Boxplot of the accuracy rate over the 10
fold cross-validation for each model (columns), loss function (rows),
intensity (x axis) and spatial normalization (color).}

\end{figure}

The main objective of this work was to test whether the typical
preprocessing in neuroimaging was needed in a CNN-DL environment. To do
so, we used images with and without applying spatial and/or intensity
normalization. By looking at the results, summarized at
Figure~\ref{fig-boxplot}, the overall conclusion is that intensity
normalization is key for CNNs to learn patterns related to PKS in our
dataset, and it has a significant effect, being the most effective the
integral normalization. There are many examples of the benefits of
intensity normalization in FP-CIT and other nuclear imaging modalities
when using traditional machine learning, e.g.~Support Vector Machines or
Principal Component Analysis (I. A. Illán et al., n.d.; F. Segovia et
al. 2012; Diego Salas-Gonzalez et al. 2009; Górriz et al. 2011).
However, while the performance gain was obvious in these cases, the
difference between normalization strategies was subtle. In our CNN,
integral normalization leads to larger performance (see ALEXNET3D and
LENET53D with the logcosh function), whereas the impact of the
normalization to the maximum is significantly smaller. This paves the
way to a more thorough analysis of intensity normalization strategies in
deep learning for neuroimaging using newer strategies, for instance,
those based on heavy-tailed distributions (Diego Salas-Gonzalez et al.
2013) or Standard Uptake Values (SUV) when the injected dose and patient
weight is available.

As for the spatial normalization, is utility could be argued. In
shallower networks such as LENET53D, it plays a more significant role,
especially when the loss function is cross-entropy. However, when using
the ALEXNET3D model, the influence of the spatial normalization is not
significant at all, neither altering the final performance nor the speed
of convergence. This could mainly be due to the CNNs accounting for
differences in alignment, rotation, scale or others. Therefore, we might
hypothesize that the deeper the network is, the less the spatial
normalization is needed.

\begin{figure}

{\centering \includegraphics{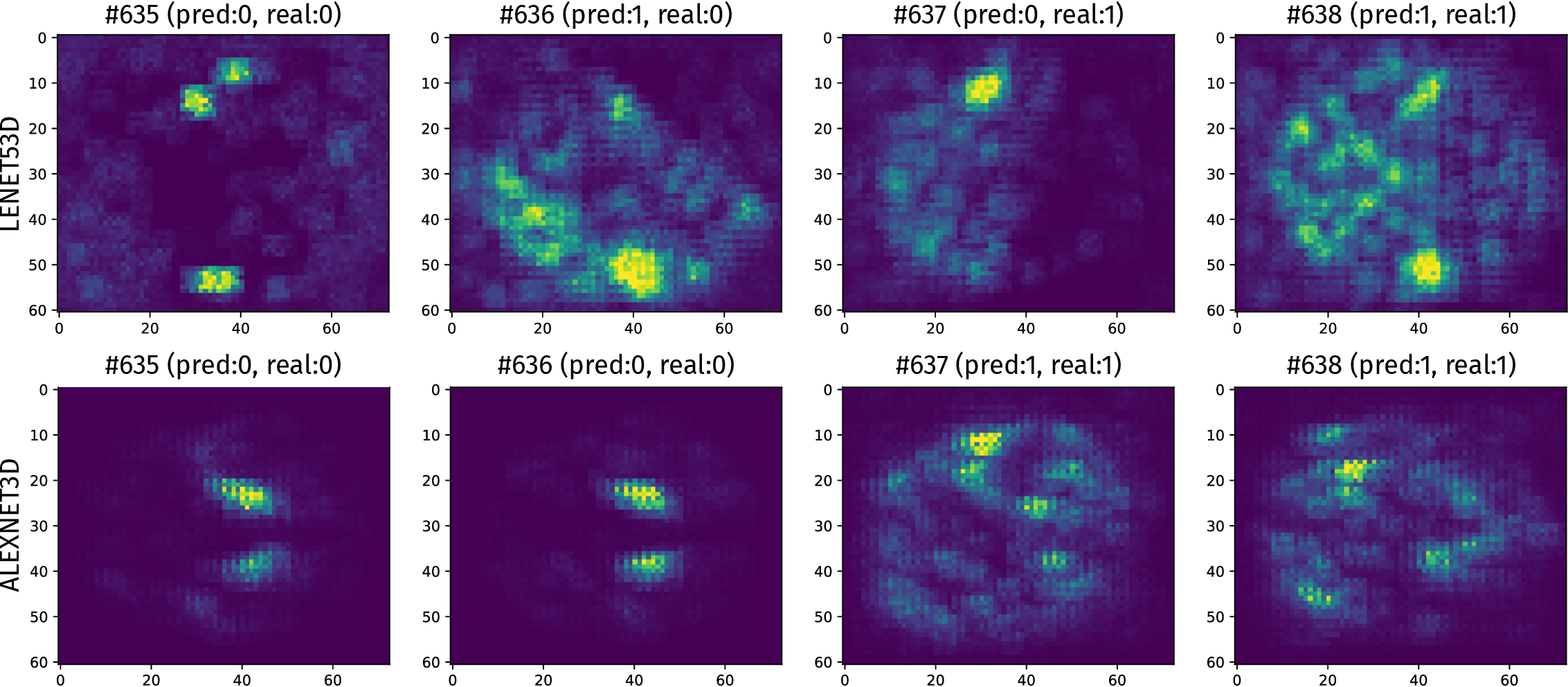}

}

\caption{\label{fig-alexnet_saliency}Saliency maps for the LENET53D
(first row) and ALEXNET3D (second row) models, in four sample images
(numbers 635, 636, 637 and 638) with integral normalization and no
spatial normalization applied (u\_int). Only the maps for the predicted
class, at neurons 0 for class control and neuron 1 for PKS, are shown.}

\end{figure}

However, recent research (Iacca et al. 2012; Pan et al. 2017) suggests
that simple metaheuristics may work better than complex metaheuristics.
In addition, the computational complexity of a deeper model is
comparable to the cost of spatial standardization in the latest
software, for example, SPM or Freesurfer. Therefore, the reader may
wonder if it is positive to build more complex models instead of using
simpler models with preprocessing. To answer this, we can see the
Figure~\ref{fig-lenet5-perf} and Figure~\ref{fig-alexnet3d-perf}.
LENET53D is a much smaller network than ALEXNET3D. However, in both
cases the performance is similar if not higher in the case of no
normalization. LENET53D achieves an accuracy of 0.922 when using
standard images while ALEXNET3D achieves an accuracy of 0.941 without
normalization. We can therefore say that, in our case, the combination
of a simpler model with spatial normalization was worse than a more
complex model without it. Then, many neuroimaging tools are shifting
towards extracting features in the image space, such as Freesurfer's
cortex thickness, volumes, etc. Furthermore, there is the case of
scalability and deployment: a Computer-Aided Diagnostics system trained
in non-standard images will always be easier to use in practice -new
acquired images can be fed directly into the system- and can potentially
be retrained without the need for spatial normalization, making it easy
to apply in real-world environments.

The saliency maps for the predicted class are depicted at
Figure~\ref{fig-alexnet_saliency}. These highlight the regions that had
more influence in the assigned class, projecting the maximum values in
the axial direction for simplicity. By comparing the LENET53D and the
ALEXNET3D architectures, the issues reported at Ref.~ are patent. Both
networks focus on elliptical features in the original images for class 0
(control). However, where the ALEXNET3D correctly finds the relative
positions of the two striata, the LENET53D is easily confused by other
features, such as the cobalt striatal phantom present in some of the
images (e.g.~635 and 637), predicting class 0 whenever it finds that
kind of pattern. Class 1 -PKS-, looks for intensity patterns in a more
distributed manner, especially weighting for relative intensity between
back and front lobes (in the LENET53D) or ration between striatum and
other regions (in ALEXNET3D). This is coherent with existing literature
on the topic (D. Salas-Gonzalez et al. 2012; Saxena et al. 1998; F.
Martínez-Murcia et al. 2014), where the control class is well defined by
intensities highly concentrated at the striatum, and the binding ratio
-ratio between intensities at the striata and non-specific areas- is
frequently used as a discrimination measure between PKS and other
etiologies (D. Salas-Gonzalez et al. 2012).

The fact that a CNN model can learn neuroimaging patterns without the
need for nonlinear registration or complex diffeomorphisms completely
changes the paradigm. Many state of the art analyses, strongly
influenced by voxelwise methodologies, require spatial normalization to
work properly. Non-rigid registration and elastic deformations to
standard templates applied in widely-known software packages eliminate
local differences via deformations which could, potentially, alter
structural and functional patterns linked to certain disorders (Keller
and Roberts 2008). With the use of CNNs, this step is required no more.
On the other hand, intensity normalization is applied globally, just
altering the intensity of the images, and introducing less variance. The
intensity normalization methodology introduced in this work is fast and
easy to apply, and produces significant benefits in the classification
of functional imaging.

Still, the main scope of this work is only to assess which preprocessing
steps are necessary in neuroimaging analysis using CNNs. Newer and more
advanced visualization techniques are still needed to identify where the
learned patterns are located and assign meaning to those, but saliency
maps of the best models are a good option for visualizing key regions in
a given class. These can be individually applied to each test image,
producing maps that precisely locate the differences on the original
image space, which support the utility of the positional invariance
introduced in ALEXNET3D. In the future, newest capsule networks (Sabour,
Frosst, and Hinton 2017) could potentially ensure coherence in the
relative position of low and higher level patterns, eliminating the need
of deeper networks. In summary, these techniques pave the way for newer
neuroimaging techniques that complement the state of the art,
potentially bearing significant advances to neuroscience.

\hypertarget{sec-conclussion}{%
\section{Conclusions}\label{sec-conclussion}}

In this work we have tested different Convolutional Neural Network (CNN)
models with the aim to test different preprocessing steps, and see if
they are relevant in CNN analysis. We have used the FP-CIT dataset
belonging to the Parkinson's Progression Markers Initiative,
preprocessed with many combinations of spatial and intensity
normalization. Different three-dimensional versions of well established
architectures, such as the LENET-5 and ALEXNET, have been used, in
combination with novel activation functions recently proposed for
self-normalizing neural networks. Different evaluation parameters and
visualization techniques have been used in order to assess the quality
of the models trained with each preprocessing pipeline. The results of
these tests show that the more complex model, the ALEXNET3D, produces
very accurate predictions, achieving a cross-validated accuracy up to
94.1\% with an area under the ROC curve of 0.984 on images with no
spatial normalization applied. This performance is even higher than the
one obtained with spatially normalized images, showing that this model
can effectively account for spatial differences. In terms of
interpretability, the features learned by the LENET53D model produced
poorer results, showing that higher abstractions -a higher number of
layers- are needed in order to provide positional invariance of the
relevant imaging features and a more generalizable diagnosis tool.
Intensity normalization, however, was proven to be very influential in
the performance of the model and must be carefully accounted for. The
visualization of the saliency maps shows that the patterns learned in
this model match those that can be found in the literature for FP-CIT
imaging, which supports the utility of this new methodology.

\hypertarget{acknowledgments}{%
\section*{Acknowledgments}\label{acknowledgments}}

This work was partly supported by the MINECO/ FEDER under the
TEC2015-64718-R project, the Consejería de Economía, Innovación, Ciencia
y Empleo (Junta de Andalucía, Spain) under the Excellence Project
P11-TIC- 7103 and the Salvador de Madariaga Mobility Grants 2017.

\hypertarget{references}{%
\section*{References}\label{references}}
\addcontentsline{toc}{section}{References}

\hypertarget{refs}{}
\begin{CSLReferences}{1}{0}
\leavevmode\vadjust pre{\hypertarget{ref-tensorflow2015-whitepaper}{}}%
Abadi, Martín, Ashish Agarwal, Paul Barham, Eugene Brevdo, Zhifeng Chen,
Craig Citro, Greg S. Corrado, et al. 2015. {``{TensorFlow}: Large-Scale
Machine Learning on Heterogeneous Systems.''}
\url{http://tensorflow.org/}.

\leavevmode\vadjust pre{\hypertarget{ref-Acharya2017}{}}%
Acharya, U. Rajendra, Shu Lih Oh, Yuki Hagiwara, Jen Hong Tan, and
Hojjat Adeli. 2017. {``Deep Convolutional Neural Network for the
Automated Detection and Diagnosis of Seizure Using {EEG} Signals.''}
\emph{Computers in Biology and Medicine}, September.
\url{https://doi.org/10.1016/j.compbiomed.2017.09.017}.

\leavevmode\vadjust pre{\hypertarget{ref-Alipanahi2015}{}}%
Alipanahi, Babak, Andrew Delong, Matthew T Weirauch, and Brendan J Frey.
2015. {``Predicting the Sequence Specificities of DNA-and RNA-Binding
Proteins by Deep Learning.''} \emph{Nature Biotechnology} 33 (8):
831--38.

\leavevmode\vadjust pre{\hypertarget{ref-Boureau2010}{}}%
Boureau, Y-Lan, Jean Ponce, and Yann LeCun. 2010. {``A Theoretical
Analysis of Feature Pooling in Visual Recognition.''} In
\emph{Proceedings of the 27th International Conference on Machine
Learning (ICML-10)}, 111--18.

\leavevmode\vadjust pre{\hypertarget{ref-Cha2017}{}}%
Cha, Young-Jin, Wooram Choi, and Oral Büyüköztürk. 2017. {``Deep
Learning-Based Crack Damage Detection Using Convolutional Neural
Networks.''} \emph{Computer-Aided Civil and Infrastructure Engineering}
32 (5): 361--78.

\leavevmode\vadjust pre{\hypertarget{ref-Ciresan2011}{}}%
Ciresan, Dan C, Ueli Meier, Jonathan Masci, Luca Maria Gambardella, and
Jürgen Schmidhuber. 2011. {``Flexible, High Performance Convolutional
Neural Networks for Image Classification.''} In \emph{IJCAI
Proceedings-International Joint Conference on Artificial Intelligence},
22:1237. 1. Barcelona, Spain.

\leavevmode\vadjust pre{\hypertarget{ref-spm_book}{}}%
Friston, K. J., J. Ashburner, S. J. Kiebel, T. E. Nichols, and W. D.
Penny. 2007. \emph{Statistical Parametric Mapping: The Analysis of
Functional Brain Images}. Academic Press.

\leavevmode\vadjust pre{\hypertarget{ref-Garrido2016}{}}%
Garrido, Jesús A, Niceto R Luque, Silvia Tolu, and Egidio D'Angelo.
2016. {``Oscillation-Driven Spike-Timing Dependent Plasticity Allows
Multiple Overlapping Pattern Recognition in Inhibitory Interneuron
Networks.''} \emph{International Journal of Neural Systems} 26 (05):
1650020.

\leavevmode\vadjust pre{\hypertarget{ref-Gawehn2016}{}}%
Gawehn, Erik, Jan A Hiss, and Gisbert Schneider. 2016. {``Deep Learning
in Drug Discovery.''} \emph{Molecular Informatics} 35 (1): 3--14.

\leavevmode\vadjust pre{\hypertarget{ref-Gorriz2017}{}}%
Gorriz, Juan M, Javier Ramirez, John Suckling, IA Illan, Andres Ortiz,
Francisco J Martinez, Fermin Segovia, Diego Salas-Gonzalez, and Shuihua
Wang. 2017. {``Case-Based Statistical Learning: A Non Parametric
Implementation with a Conditional-Error Rate SVM.''} \emph{IEEE Access}.

\leavevmode\vadjust pre{\hypertarget{ref-Gorriz2010}{}}%
Górriz, J. M., F. Segovia, J. Ramírez, A. Lassl, and D. Salas-Gonzalez.
2011. {``GMM Based {SPECT} Image Classification for the Diagnosis of
{Alzheimer}'s Disease.''} \emph{Applied Soft Computing} 11 (2):
2313--25. \url{https://doi.org/10.1016/j.asoc.2010.08.012}.

\leavevmode\vadjust pre{\hypertarget{ref-Greenspan2016}{}}%
Greenspan, Hayit, Bram van Ginneken, and Ronald M Summers. 2016.
{``Guest Editorial Deep Learning in Medical Imaging: Overview and Future
Promise of an Exciting New Technique.''} \emph{IEEE Transactions on
Medical Imaging} 35 (5): 1153--59.

\leavevmode\vadjust pre{\hypertarget{ref-Hanley1982}{}}%
Hanley, James A, and Barbara J McNeil. 1982. {``The Meaning and Use of
the Area Under a Receiver Operating Characteristic (ROC) Curve.''}
\emph{Radiology} 143 (1): 29--36.

\leavevmode\vadjust pre{\hypertarget{ref-Hinton2012}{}}%
Hinton, Geoffrey, Li Deng, Dong Yu, George E Dahl, Abdel-rahman Mohamed,
Navdeep Jaitly, Andrew Senior, et al. 2012. {``Deep Neural Networks for
Acoustic Modeling in Speech Recognition: The Shared Views of Four
Research Groups.''} \emph{IEEE Signal Processing Magazine} 29 (6):
82--97.

\leavevmode\vadjust pre{\hypertarget{ref-Hirschauer2015}{}}%
Hirschauer, Thomas J, Hojjat Adeli, and John A Buford. 2015.
{``Computer-Aided Diagnosis of Parkinson's Disease Using Enhanced
Probabilistic Neural Network.''} \emph{Journal of Medical Systems} 39
(11): 179.

\leavevmode\vadjust pre{\hypertarget{ref-Iacca2012}{}}%
Iacca, Giovanni, Ferrante Neri, Ernesto Mininno, Yew-Soon Ong, and
Meng-Hiot Lim. 2012. {``Ockham's Razor in Memetic Computing: Three Stage
Optimal Memetic Exploration.''} \emph{Information Sciences} 188: 17--43.

\leavevmode\vadjust pre{\hypertarget{ref-Illan2012}{}}%
Illán, I. A.a, J. M.a Górriz, J.a Ramírez, F.a Segovia, J. M.b
Jiménez-Hoyuela, and S. J.b Ortega Lozano. 2012. {``Automatic Assistance
to {Parkinson}'s Disease Diagnosis in {DaTSCAN} {SPECT} Imaging.''}
\emph{Medical Physics} 39 (10): 5971--80.

\leavevmode\vadjust pre{\hypertarget{ref-Illan2011}{}}%
Illán, I. A., J. M. Górriz, J. Ramírez, D. Salas-Gonzalez, M. M. López,
F. Segovia, R. Chaves, M. Gómez-Rio, and C. G. Puntonet. n.d.
{``{18F-FDG PET} Imaging Analysis for Computer Aided {Alzheimer's}
Diagnosis.''}

\leavevmode\vadjust pre{\hypertarget{ref-Inititative2010}{}}%
Initiative, The Parkinson Progression Markers. 2010. \emph{PPMI. Imaging
Technical Operations Manual}. 2nd ed.

\leavevmode\vadjust pre{\hypertarget{ref-Ishii2001}{}}%
Ishii, Kazunari, Frode Willoch, Satoshi Minoshima, Alexander Drzezga,
Edward P Ficaro, Donna J Cross, David E Kuhl, and Markus Schwaiger.
2001. {``Statistical Brain Mapping of {18F}-{FDG} {PET} in Alzheimer's
Disease: Validation of Anatomic Standardization for Atrophied Brains.''}
\emph{Journal of Nuclear Medicine} 42 (4): 548--57.

\leavevmode\vadjust pre{\hypertarget{ref-Keller2008}{}}%
Keller, Simon Sean, and Neil Roberts. 2008. {``Voxel-Based Morphometry
of Temporal Lobe Epilepsy: An Introduction and Review of the
Literature.''} \emph{Epilepsia} 49 (5): 741--57.

\leavevmode\vadjust pre{\hypertarget{ref-Kim2004}{}}%
Kim, Kyungsun, Harksoo Kim, and Jungyun Seo. 2004. {``A Neural Network
Model with Feature Selection for Korean Speech Act Classification.''}
\emph{International Journal of Neural Systems} 14 (06): 407--14.

\leavevmode\vadjust pre{\hypertarget{ref-Klambauer2017}{}}%
Klambauer, Günter, Thomas Unterthiner, Andreas Mayr, and Sepp
Hochreiter. 2017. {``Self-Normalizing Neural Networks.''} \emph{arXiv
Preprint arXiv:1706.02515}.

\leavevmode\vadjust pre{\hypertarget{ref-Kohavi1995}{}}%
Kohavi, Ron, and George H. John. 1995. {``Wrappers for Feature Subset
Selection.''} \emph{AIJ Special Issue on Relevance}.

\leavevmode\vadjust pre{\hypertarget{ref-Kotikalapudi2017}{}}%
Kotikalapudi, Raghavendra, and contributors. 2017. {``Keras-Vis.''}
\url{https://github.com/raghakot/keras-vis}; GitHub.

\leavevmode\vadjust pre{\hypertarget{ref-Koziarski2017}{}}%
Koziarski, Michał, and Bogusław Cyganek. 2017. {``Image Recognition with
Deep Neural Networks in Presence of Noise--Dealing with and Taking
Advantage of Distortions.''} \emph{Integrated Computer-Aided
Engineering} 24 (4): 337--49.

\leavevmode\vadjust pre{\hypertarget{ref-Krizhevsky2012}{}}%
Krizhevsky, Alex, Ilya Sutskever, and Geoffrey E Hinton. 2012.
{``Imagenet Classification with Deep Convolutional Neural Networks.''}
In \emph{Advances in Neural Information Processing Systems}, 1097--1105.

\leavevmode\vadjust pre{\hypertarget{ref-LeCun2015}{}}%
LeCun, Yann, Yoshua Bengio, and Geoffrey Hinton. 2015. {``Deep
Learning.''} \emph{Nature} 521 (7553): 436--44.
\url{https://doi.org/10.1038/nature14539}.

\leavevmode\vadjust pre{\hypertarget{ref-LeCun1998}{}}%
LeCun, Yann, Léon Bottou, Yoshua Bengio, and Patrick Haffner. 1998.
{``Gradient-Based Learning Applied to Document Recognition.''}
\emph{Proceedings of the IEEE} 86 (11): 2278--2324.

\leavevmode\vadjust pre{\hypertarget{ref-Liao2013}{}}%
Liao, Shu, Yaozong Gao, Aytekin Oto, and Dinggang Shen. 2013.
{``Representation Learning: A Unified Deep Learning Framework for
Automatic Prostate MR Segmentation.''} In \emph{International Conference
on Medical Image Computing and Computer-Assisted Intervention}, 254--61.
Springer.

\leavevmode\vadjust pre{\hypertarget{ref-Lin2017}{}}%
Lin, Yi-zhou, Zhen-hua Nie, and Hong-wei Ma. 2017. {``Structural Damage
Detection with Automatic Feature-Extraction Through Deep Learning.''}
\emph{Computer-Aided Civil and Infrastructure Engineering} 32 (12):
1025--46.

\leavevmode\vadjust pre{\hypertarget{ref-Lopez2011}{}}%
López, M., J. Ramírez, J. M. Górriz, I. Álvarez, D. Salas-Gonzalez, F.
Segovia, R. Chaves, P. Padilla, and M. Gómez-Río. 2011. {``Principal
Component Analysis-Based Techniques and Supervised Classification
Schemes for the Early Detection of {Alzheimer's} Disease.''}
\emph{Neurocomputing} 74 (8): 1260--71.
\url{https://doi.org/10.1016/j.neucom.2010.06.025}.

\leavevmode\vadjust pre{\hypertarget{ref-Martinez-Murcia2016book}{}}%
Martinez-Murcia, F. J., J. Górriz, and J. Ramírez. 2016. {``Computer
Aided Diagnosis in Neuroimaging.''} In \emph{Computer-Aided Technologies
- Applications in Engineering and Medicine}, edited by Razvan Udroiu,
1st ed., 137--60. InTech. \url{https://doi.org/10.5772/64980}.

\leavevmode\vadjust pre{\hypertarget{ref-Martinez-Murcia2017a}{}}%
Martinez-Murcia, Francisco Jesús, Andres Ortiz, Juan Manuel Górriz,
Javier Ramírez, Fermin Segovia, Diego Salas-Gonzalez, Diego
Castillo-Barnes, and Ignacio A. Illán. 2017. {``A {3D} Convolutional
Neural Network Approach for the Diagnosis of {Parkinson}'s Disease.''}
In \emph{Natural and Artificial Computation for Biomedicine and
Neuroscience}, 324--33. Springer International Publishing.
\url{https://doi.org/10.1007/978-3-319-59740-9_32}.

\leavevmode\vadjust pre{\hypertarget{ref-Martinez-Murcia2012}{}}%
Martínez-Murcia, F. J., J. M. Górriz, J. Ramírez, C. G. Puntonet, and D.
Salas-González. 2012. {``Computer Aided Diagnosis Tool for {Alzheimer's}
Disease Based on {Mann-Whitney-Wilcoxon U}-Test.''} \emph{Expert Systems
with Applications} 39 (10): 9676--85.
\url{https://doi.org/10.1016/j.eswa.2012.02.153}.

\leavevmode\vadjust pre{\hypertarget{ref-martinez2014parametrization}{}}%
Martínez-Murcia, FJ, JM Górriz, J Ramírez, M Moreno-Caballero, M
Gómez-Río, Parkinson's Progression Markers Initiative, et al. 2014.
{``Parametrization of Textural Patterns in {123I}-Ioflupane Imaging for
the Automatic Detection of Parkinsonism.''} \emph{Medical Physics} 41
(1): 012502.

\leavevmode\vadjust pre{\hypertarget{ref-Martin-Lopez2017}{}}%
Martín-López, David, Diego Jiménez-Jiménez, Lidia Cabañés-Martínez,
Richard P Selway, Antonio Valentín, and Gonzalo Alarcón. 2017. {``The
Role of Thalamus Versus Cortex in Epilepsy: Evidence from Human Ictal
Centromedian Recordings in Patients Assessed for Deep Brain
Stimulation.''} \emph{International Journal of Neural Systems} 27 (07):
1750010.

\leavevmode\vadjust pre{\hypertarget{ref-Martino2013}{}}%
Martino, María Elena, Juan Guzmán de Villoria, María Lacalle-Aurioles,
Javier Olazarán, Isabel Cruz, Eloisa Navarro, Verónica García-Vázquez,
José Luis Carreras, and Manuel Desco. 2013. {``Comparison of Different
Methods of Spatial Normalization of {FDG}-{PET} Brain Images in the
Voxel-Wise Analysis of {MCI} Patients and Controls.''} \emph{Annals of
Nuclear Medicine} 27 (7): 600--609.
\url{https://doi.org/10.1007/s12149-013-0723-7}.

\leavevmode\vadjust pre{\hypertarget{ref-Mazziotta2001}{}}%
Mazziotta, John, Arthur Toga, Alan Evans, Peter Fox, Jack Lancaster,
Karl Zilles, Roger Woods, et al. 2001. {``A Probabilistic Atlas and
Reference System for the Human Brain: International Consortium for Brain
Mapping (ICBM).''} \emph{Philosophical Transactions of the Royal Society
of London B: Biological Sciences} 356 (1412): 1293--1322.

\leavevmode\vadjust pre{\hypertarget{ref-Morabito2017}{}}%
Morabito, Francesco Carlo, Maurizio Campolo, Nadia Mammone, Mario
Versaci, Silvana Franceschetti, Fabrizio Tagliavini, Vito Sofia, et al.
2017. {``Deep Learning Representation from Electroencephalography of
Early-Stage Creutzfeldt-Jakob Disease and Features for Differentiation
from Rapidly Progressive Dementia.''} \emph{International Journal of
Neural Systems} 27 (02): 1650039.

\leavevmode\vadjust pre{\hypertarget{ref-Olson2013}{}}%
Olson, Larry D, and M Scott Perry. 2013. {``Localization of Epileptic
Foci Using Multimodality Neuroimaging.''} \emph{International Journal of
Neural Systems} 23 (01): 1230001.

\leavevmode\vadjust pre{\hypertarget{ref-Ortega-Zamorano2017}{}}%
Ortega-Zamorano, Francisco, José M Jerez, Iván Gómez, and Leonardo
Franco. 2017. {``Layer Multiplexing FPGA Implementation for Deep
Back-Propagation Learning.''} \emph{Integrated Computer-Aided
Engineering} 24 (2): 171--85.

\leavevmode\vadjust pre{\hypertarget{ref-Ortiz2016}{}}%
Ortiz, Andrés, Francisco J Martínez-Murcia, María J García-Tarifa,
Francisco Lozano, Juan M Górriz, and Javier Ramírez. 2016. {``Automated
Diagnosis of Parkinsonian Syndromes by Deep Sparse Filtering-Based
Features.''} In \emph{Innovation in Medicine and Healthcare 2016},
249--58. Springer.

\leavevmode\vadjust pre{\hypertarget{ref-Pan2017}{}}%
Pan, Linqiang, Gheorghe Păun, Gexiang Zhang, and Ferrante Neri. 2017.
{``Spiking Neural p Systems with Communication on Request.''}
\emph{International Journal of Neural Systems} 27 (08): 1750042.

\leavevmode\vadjust pre{\hypertarget{ref-Payan2015}{}}%
Payan, Adrien, and Giovanni Montana. 2015. {``Predicting {Alzheimer's}
Disease: A Neuroimaging Study with {3D} Convolutional Neural
Networks.''} \emph{arXiv Preprint arXiv:1502.02506}.

\leavevmode\vadjust pre{\hypertarget{ref-Rafiei2015}{}}%
Rafiei, Mohammad Hossein, and Hojjat Adeli. 2015. {``A Novel Machine
Learning Model for Estimation of Sale Prices of Real Estate Units.''}
\emph{Journal of Construction Engineering and Management} 142 (2):
04015066.

\leavevmode\vadjust pre{\hypertarget{ref-Rafiei2017a}{}}%
---------. 2017. {``A Novel Machine Learning-Based Algorithm to Detect
Damage in High-Rise Building Structures.''} \emph{The Structural Design
of Tall and Special Buildings} 26 (18).

\leavevmode\vadjust pre{\hypertarget{ref-Rafiei2018}{}}%
---------. 2018. {``A Novel Unsupervised Deep Learning Model for Global
and Local Health Condition Assessment of Structures.''}
\emph{Engineering Structures} 156: 598--607.

\leavevmode\vadjust pre{\hypertarget{ref-Rafiei2017}{}}%
Rafiei, Mohammad Hossein, Waleed H Khushefati, Ramazan Demirboga, and
Hojjat Adeli. 2017. {``Supervised Deep Restricted Boltzmann Machine for
Estimation of Concrete.''} \emph{ACI Materials Journal} 114 (2).

\leavevmode\vadjust pre{\hypertarget{ref-Ramirez2010}{}}%
Ramírez, J., J. M. Górriz, F. Segovia, R. Chaves, D. Salas-Gonzalez, M.
López, I. Álvarez, and P. Padilla. 2010. {``Computer Aided Diagnosis
System for the {Alzheimer's} Disease Based on Partial Least Squares and
Random Forest {SPECT} Image Classification.''} \emph{Neuroscience
Letters} 472 (2): 99--103.

\leavevmode\vadjust pre{\hypertarget{ref-Reig2007}{}}%
Reig, S., M. Penedo, J. D. Gispert, J. Pascau, J. Sánchez-González, P.
García-Barreno, and M. Desco. 2007. {``Impact of Ventricular Enlargement
on the Measurement of Metabolic Activity in Spatially Normalized
{PET}.''} \emph{{NeuroImage}} 35 (2): 748--58.
\url{https://doi.org/10.1016/j.neuroimage.2006.12.015}.

\leavevmode\vadjust pre{\hypertarget{ref-Reuter2010}{}}%
Reuter, Martin, H Diana Rosas, and Bruce Fischl. 2010. {``Highly
Accurate Inverse Consistent Registration: A Robust Approach.''}
\emph{Neuroimage} 53 (4): 1181--96.

\leavevmode\vadjust pre{\hypertarget{ref-Sabour2017}{}}%
Sabour, Sara, Nicholas Frosst, and Geoffrey E Hinton. 2017. {``Dynamic
Routing Between Capsules.''} In \emph{Advances in Neural Information
Processing Systems}, 3857--67.

\leavevmode\vadjust pre{\hypertarget{ref-salas2012intensity}{}}%
Salas-Gonzalez, D, JM Gorriz, J Ramirez, FJ Martinez, R Chaves, F
Segovia, and IA Illan. 2012. {``Intensity Normalization of {FP}-CIT
{SPECT} in Patients with Parkinsonism Using the \(\alpha\)-Stable
Distribution.''} In \emph{Nuclear Science Symposium and Medical Imaging
Conference (NSS/MIC), 2012 IEEE}, 3944--46. IEEE.

\leavevmode\vadjust pre{\hypertarget{ref-Salas-Gonzalez2013}{}}%
Salas-Gonzalez, Diego, Juan M. Górriz, Javier Ramírez, Ignacio A. Illán,
and Elmar W. Lang. 2013. {``Linear Intensity Normalization of {FP}-{CIT}
{SPECT} Brain Images Using the \(\alpha\)-Stable Distribution.''}
\emph{{NeuroImage}} 65 (January): 449--55.
\url{https://doi.org/10.1016/j.neuroimage.2012.10.005}.

\leavevmode\vadjust pre{\hypertarget{ref-Salas-Gonzalez2009}{}}%
Salas-Gonzalez, Diego, Juan M. Górriz, Javier Ramírez, Miriam López,
Ignacio A. Illan, Fermín Segovia, Carlos G. Puntonet, and Manuel
Gómez-Río. 2009. {``Analysis of {SPECT} Brain Images for the Diagnosis
of {Alzheimer's} Disease Using Moments and Support Vector Machines.''}
\emph{Neuroscience Letters} 461 (September): 60--64.

\leavevmode\vadjust pre{\hypertarget{ref-Saxena1998}{}}%
Saxena, P., D. G. Pavel, J. C. Quintana, and B. Horwitz. 1998. {``An
Automatic Threshold-Based Scaling Method for Enhancing the Usefulness of
{Tc-HMPAO SPECT} in the Diagnosis of {Alzheimer's} Disease.''} In
\emph{Medical Image Computing and Computer-Assisted Intervention -
MICCAI}, 1496:623--30. Lecture Notes in Computer Science. Springer.

\leavevmode\vadjust pre{\hypertarget{ref-Schmidhuber2015}{}}%
Schmidhuber, Jürgen. 2015. {``Deep Learning in Neural Networks: An
Overview.''} \emph{Neural Networks} 61: 85--117.

\leavevmode\vadjust pre{\hypertarget{ref-Segovia2016}{}}%
Segovia, Fermín, Marcelo García-Pérez, Juan Manuel Górriz, Javier
Ramírez, and Francisco Jesús Martínez-Murcia. 2016. {``Assisting the
Diagnosis of Neurodegenerative Disorders Using Principal Component
Analysis and TensorFlow.''} In \emph{International Conference on
EUropean Transnational Education}, 43--52. Springer.

\leavevmode\vadjust pre{\hypertarget{ref-Segovia2012a}{}}%
Segovia, Fermín, Juan Manuel Górriz, Javier Ramírez, Rosa Chaves, and I
Álvarez Illán. 2012. {``Automatic Differentiation Between Controls and
{Parkinson}'s Disease {DaTSCAN} Images Using a Partial Least Squares
Scheme and the Fisher Discriminant Ratio.''} In \emph{KES}, 2241--50.

\leavevmode\vadjust pre{\hypertarget{ref-Segovia2012}{}}%
Segovia, F., J. M. Górriz, J. Ramírez, I. Álvarez, J. M.
Jiménez-Hoyuela, and S. J. Ortega. 2012. {``Improved Parkinsonism
Diagnosis Using a Partial Least Squares Based Approach.''} \emph{Medical
Physics} 39 (7): 4395--4403.

\leavevmode\vadjust pre{\hypertarget{ref-Simonyan2013}{}}%
Simonyan, Karen, Andrea Vedaldi, and Andrew Zisserman. 2013. {``Deep
Inside Convolutional Networks: Visualising Image Classification Models
and Saliency Maps.''} \emph{arXiv Preprint arXiv:1312.6034}.

\leavevmode\vadjust pre{\hypertarget{ref-Springenberg2014}{}}%
Springenberg, Jost Tobias, Alexey Dosovitskiy, Thomas Brox, and Martin
Riedmiller. 2014. {``Striving for Simplicity: The All Convolutional
Net.''} \emph{arXiv Preprint arXiv:1412.6806}.

\leavevmode\vadjust pre{\hypertarget{ref-Xu2014}{}}%
Xu, Yan, Tao Mo, Qiwei Feng, Peilin Zhong, Maode Lai, I Eric, and Chao
Chang. 2014. {``Deep Learning of Feature Representation with Multiple
Instance Learning for Medical Image Analysis.''} In \emph{Acoustics,
Speech and Signal Processing (ICASSP), 2014 IEEE International
Conference on}, 1626--30. IEEE.

\leavevmode\vadjust pre{\hypertarget{ref-Yuvaraj2016}{}}%
Yuvaraj, R, M Murugappan, U Rajendra Acharya, Hojjat Adeli, Norlinah
Mohamed Ibrahim, and Edgar Mesquita. 2016. {``Brain Functional
Connectivity Patterns for Emotional State Classification in Parkinson's
Disease Patients Without Dementia.''} \emph{Behavioural Brain Research}
298: 248--60.

\leavevmode\vadjust pre{\hypertarget{ref-Zhang2017}{}}%
Zhang, Allen, Kelvin CP Wang, Baoxian Li, Enhui Yang, Xianxing Dai, Yi
Peng, Yue Fei, Yang Liu, Joshua Q Li, and Cheng Chen. 2017. {``Automated
Pixel-Level Pavement Crack Detection on 3D Asphalt Surfaces Using a
Deep-Learning Network.''} \emph{Computer-Aided Civil and Infrastructure
Engineering} 32 (10): 805--19.

\end{CSLReferences}

\end{document}